\documentclass[lettersize,journal]{IEEEtran}
\usepackage{amsmath,amsfonts}
\usepackage{amsthm}
\usepackage{multirow}
\usepackage[]{threeparttable}
\usepackage{xspace}
\usepackage{bm}
\usepackage{booktabs}
\usepackage{algorithmic}
\usepackage{algorithm}
\usepackage{array}
\usepackage[caption=false,font=normalsize,labelfont=sf,textfont=sf]{subfig}
\usepackage{textcomp}
\usepackage{stfloats}
\usepackage{url}
\usepackage{verbatim}
\usepackage{hyperref}
\usepackage{graphicx}
\usepackage{cite}
\usepackage{color}
\usepackage{colortbl}
\allowdisplaybreaks[4]

\definecolor{tabGray}{gray}{0.96}

\DeclareMathOperator*{\rvec}{\mathrm{vec}}
\theoremstyle{remark}

\begin{document}

\title{Natural Gradient Gaussian Approximation Filter on Lie Groups \\ for Robot State Estimation}

\author{Tianyi Zhang, Wenhan Cao, Chang Liu, Yao Lyu, and Shengbo Eben Li
\thanks{This study is supported by Tsinghua-Efort Joint Research Center for EAI Computation and Perception.}
\thanks{Tianyi Zhang, Wenhan Cao and Yao Lyu are with the School of Vehicle and Mobility, Tsinghua University, Beijing, China (e-mail: \{zhangtia24, cwh19,  \}@mails.tsinghua.edu.cn, 
lyo.tobias@foxmail.com). Tianyi Zhang and Wenhan Cao contributed equally to this work.}
\thanks{Chang Liu is with the College of Engineering, Peking University, Beijing, China (e-mail: changliucoe@pku.edu.cn).
}
\thanks{Shengbo Eben Li is with the School of Vehicle and Mobility and College
of Artificial Intelligence, Tsinghua University, Beijing, China (e-mail: lish04@gmail.com).}

\thanks{Corresponding Author: Shengbo Eben Li}
}


\maketitle

\begin{abstract}
Accurate state estimation for robotic systems evolving on Lie group manifolds, such as legged robots, is a prerequisite for achieving agile control. However, this task is challenged by nonlinear observation models defined on curved manifolds, where existing filters rely on local linearization in the tangent space to handle such nonlinearity, leading to accumulated estimation errors. To address this limitation, we reformulate manifold filtering as a parameter optimization problem over a Gaussian-distributed increment variable, thereby avoiding linearization.
Under this formulation, the increment can be mapped to the Lie group through the exponential operator, where it acts multiplicatively on the prior estimate to yield the posterior state.
We further propose a natural gradient optimization scheme for solving this problem, whose iteration process leverages the Fisher information matrix of the increment variable to account for the curvature of the tangent space. This results in an iterative algorithm named the Natural Gradient Gaussian Approximation on Lie Groups (NANO-L) filter.
Leveraging the perturbation model in Lie derivative, we prove that for the invariant observation model widely adopted in robotic localization tasks, the covariance update in NANO-L admits an exact closed-form solution, eliminating the need for iterative updates thus improving computational efficiency.
Hardware experiments on a Unitree GO2 legged robot operating across different terrains demonstrate that NANO-L achieves approximately 40\% lower estimation error than commonly used filters at a comparable computational cost.
\end{abstract}



\section{Introduction}
Autonomous robots have become increasingly important in a wide range of applications, such as search and rescue, warehouse logistics, and manufacturing assembly.
To develop motion planners and feedback controllers for these tasks, accurate estimation of the robot’s states like pose and velocity is essential \cite{hartley2020contact, zhang2021pose, barfoot2024state}. 
The predominant architecture for robot state estimation relies on fusing data from onboard sensors, such as inertial measurement units (IMUs) and wheel or joint encoders, using a filter \cite{bloesch2013state, yu2023fully}. This filtering-based approach is particularly favored for its balance of accuracy, low cost, and computational efficiency, making it suitable for real-time operation on resource-constrained platforms.

\begin{figure}[!t]
\centering
\includegraphics[width=0.48\textwidth]{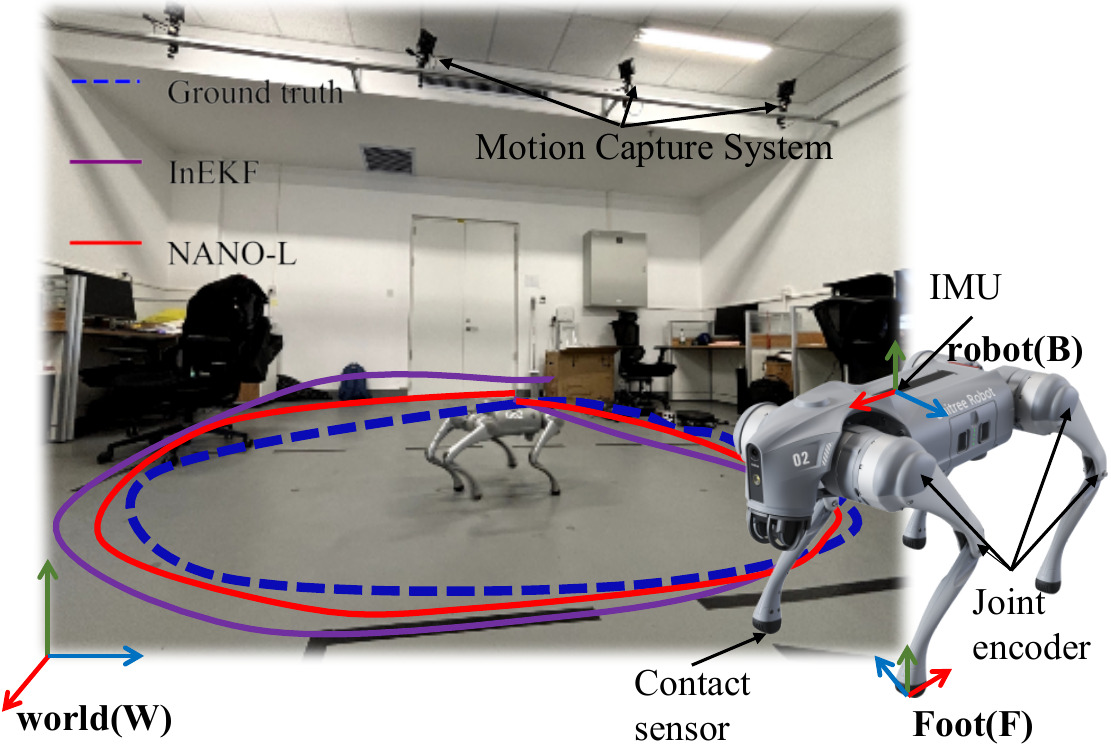}
\caption{A legged robot equipped with an IMU, joint encoders, and contact sensors is walking in an indoor environment equipped with a motion capture system. The estimated trajectory of the proposed NANO-L is closer to the ground truth than that of InEKF.}
\label{fig:intro}
\end{figure}

Classical filters such as the extended Kalman filter (EKF) \cite{bloesch2013state} and the unscented Kalman filter (UKF) \cite{bloesch2013state2} have been widely used in nonlinear robotic systems for many years. However, these methods share a fundamental limitation: they assume that the state evolves in a Euclidean vector space \cite{barrau2016invariant, brossard2020code, barfoot2024state}. In reality, the state of autonomous robots evolves on smooth manifolds known as Lie groups. For example, for the Unitree GO2 robot shown in Fig.~\ref{fig:intro}, its orientation lies on the special orthogonal group \(\mathrm{SO}(3)\), and its pose lies on the special Euclidean group \(\mathrm{SE}(3)\)\cite{zhang2017convergence,hartley2020contact,barfoot2024state}. Applying Euclidean filters to such non-Euclidean states ignores the underlying geometry, which can lead to singularities, constraint violations, and instability \cite{barrau2016invariant,hartley2020contact,zhu2022design}. This critical shortcoming has motivated extensive research into filtering algorithms that operate on manifolds.

Early approaches to manifold filtering typically focus on variations of EKF. One of the earliest attempts is the multiplicative EKF (MEKF) \cite{markley2003attitude}, which estimates attitude on the unit quaternion group via multiplicative error representations, naturally handling rotational constraints. Nevertheless, MEKF is not suitable for broader state representations due to its restriction to quaternion-based attitude states. 
A significant step toward a more general framework was the introduction of concentrated Gaussian distributions on Lie groups \cite{barfoot2014associating}.
This concept represents uncertainty as a Gaussian-distributed perturbation vector in the group's tangent space, while maintaining the mean on the manifold itself. Building on this formulation, the EKF can be applied to estimate this perturbation vector \cite{bourmaud2013discrete, bourmaud2015continuous}, which is applicable to general Lie groups.
Nonetheless, the EKF’s reliance on linearization around potentially inaccurate state estimates can amplify errors and lead to divergence \cite{barrau2016invariant,hartley2020contact}. To address this, a notable development is the Invariant EKF (InEKF) \cite{barrau2016invariant}, which shows that for systems with group-affine dynamics and invariant observations, the linearization is independent of the current estimate, leading to improved convergence and stability in numerous robotic state estimation tasks \cite{zhang2017convergence, hartley2020contact, zhu2022design, yu2023fully}. Unfortunately, the InEKF's strict requirements on system structure limit its applicability, and its accuracy is still constrained by the first-order Taylor linearization in EKF.

In parallel to the development of manifold-based EKF, recent works have also extended UKF to manifolds, capitalizing on its superior ability to handle strong nonlinearities compared to EKF \cite{julier2004unscented,cao2024nonlinear}.
Brossard et al. proposed the UKF on Manifolds (UKF-M) \cite{brossard2017unscented,brossard2020code}, which refines sigma points defined in the tangent space by applying the unscented transformation (UT) adapted for Lie groups. Its effectiveness has been demonstrated in applications such as inertial navigation and visual localization, where it achieves higher accuracy than EKF variants.
However, since sigma points are defined in the tangent space while UT is performed on the Lie group, performing one UT requires four transformations between the tangent space and the group, which introduces additional computational overhead. While recent efforts have attempted to improve efficiency by performing the UT entirely in the tangent space \cite{sjoberg2021lie, jin2024nonlinear}, this strategy depends on first-order Taylor approximations of system dynamics, thereby introducing larger linearization errors. Moreover, the UT itself can be regarded as a stochastic linear regression method, which inherently introduces certain linearization errors \cite{cao2024nonlinear}. Thus, whether based on EKF or UKF, existing manifold filters are still fundamentally tied to local approximation schemes, either taylor expansion or UT, which can be inaccurate and computationally demanding.

To address the limitations of local linearization in existing manifold filters, we propose the Natural Gradient Gaussian Approximation on Lie Groups (NANO-L) filter.
Our approach formulates state estimation on Lie groups as a direct Bayesian inference problem in the tangent space and solves it through natural gradient optimization to achieve both accuracy and efficiency. Through hardware experiments on a state estimation task for the Unitree GO2 legged robot operating across different terrains, we demonstrate that the proposed NANO-L filter achieves over a 40\% improvement in estimation accuracy on average relative to the best existing filtering algorithms, while incurring a comparable computational cost. Our main contributions are summarized as follows:

\begin{enumerate}
\item We reframe manifold filtering as an optimization problem over the parameters of a Gaussian-distributed increment variable in the tangent space and show that the optimal increment minimizes the sum of an expected likelihood term and regularization terms that penalize discrepancies in the mean, covariance, and volume between the prior and posterior distributions. Under this formulation, the optimal increment is mapped from the tangent space to the Lie group via the exponential map, where it updates the prior estimate to produce the posterior state.

\item We propose a natural gradient optimization scheme for solving this problem, where each iteration leverages the Riemannian metric of the statistical manifold to account for the geometry of the tangent space. The resulting NANO-L filter avoids the approximation errors caused by local linearization in conventional EKF-based and UKF-based manifold filtering methods, thereby achieving higher estimation accuracy.

\item We further prove that, for commonly used invariant observation models in robotics such as leg, wheel, and visual-inertial odometry, the covariance update of NANO-L admits an exact closed-form solution derived through the perturbation model in Lie derivative. This eliminates the need for iterative refinement, further improving the computational efficiency of NANO-L.
\end{enumerate}

The remainder of this paper is organized as follows. Section II introduces the preliminaries of Lie group theory that form the mathematical foundation of our formulation. Section III presents the system modeling for robot state estimation on Lie groups. Section IV details the design procedure of the proposed NANO-L filter. Section V provides numerical simulation results that validate the effectiveness of the proposed algorithm, and Section VI reports real-world experiments on a Unitree GO2 legged robot to demonstrate its practical performance. Finally, Section VII concludes the paper.

\section{Preliminaries}
This section introduces the mathematical background required for developing NANO-L algorithm. We first review the fundamentals of Lie groups, including the exponential and logarithm maps, followed by the Baker–Campbell–Hausdorff (BCH) formula. We then discuss how uncertainties can be represented on Lie groups, and conclude with two important examples frequently encountered in robotics.

\subsection{Lie Groups Theory}
\label{sec:pre_a}
A matrix Lie group, denoted \(\mathcal{G}\), is a non-commutative group consisting of \(n\times n\) matrices under matrix multiplication. It is also a smooth manifold, allowing for differential operations to be carried out on it. Each matrix Lie group corresponds to a Lie algebra \(\mathfrak{g}\) that serves as the tangent space at the group identity and captures the Lie group’s infinitesimal structure. 
The conversion between Lie groups and its Lie algebras can be achieved through matrix exponential \(\exp\) and matrix logarithm \(\log\) operations, as
\begin{equation}
\nonumber
\begin{aligned}
\exp : \mathfrak{g} \rightarrow \mathcal{G}, &\quad \log : \mathcal{G} \rightarrow \mathfrak{g},
\end{aligned}
\end{equation}
which together define a local diffeomorphism between \(\mathcal{G}\) and \(\mathfrak{g}\). Although the Lie algebra is represented by matrices of the same dimension as the group elements, it is often more convenient to work with its Euclidean space representation \(\mathbb{R}^{\dim \mathfrak{g}}\). This is achieved through the \textit{wedge} and \textit{vee} operators:
\begin{equation}
\nonumber
\begin{aligned}
(\cdot)^{\wedge}: \mathbb{R}^{\dim \mathfrak{g}} \rightarrow \mathfrak{g},& \quad (\cdot)^{\vee}: \mathfrak{g} \rightarrow \mathbb{R}^{\dim \mathfrak{g}}.
\end{aligned}
\end{equation}
For clarity, we adopt the following notation for the mapping between the Euclidean space and the group:
\begin{equation}
\label{eq.Exp_Log}
\mathrm{Exp} : \mathbb{R}^{\dim \mathfrak{g}} \rightarrow \mathcal{G}, \quad \mathrm{Log} : \mathcal{G} \rightarrow \mathbb{R}^{\dim \mathfrak{g}}.
\end{equation}
With this notation, the Euclidean space $\mathbb{R}^{\dim \mathfrak{g}}$ will be used in place of the Lie algebra \(\mathfrak{g}\) to denote the tangent space of the Lie group throughout the remainder of the paper for simplicity.

The Baker-Campbell-Hausdorff (BCH) formula plays a central role when manipulating products of Lie group elements. Consider two vectors \(\bm{x}_1, \bm{x}_2 \in \mathbb{R}^{\dim \mathfrak{g}}\) associated with different group elements \cite{barfoot2024state}. When one of the vectors is small, the BCH formula provides a useful approximation for combining them:
\begin{equation}
\label{eq.BCH}
\begin{aligned}
\mathrm{Exp} \left(\boldsymbol{x}_1\right) \mathrm{Exp}\left(\boldsymbol{x}_2\right) &\approx \begin{cases}\mathrm{Exp} (\boldsymbol{J}_{l}\left(\boldsymbol{x}_2\right)^{-1} \boldsymbol{x}_1+\boldsymbol{x}_2) & \boldsymbol{x}_1 \text { small } \\ \mathrm{Exp} (\boldsymbol{x}_1+\boldsymbol{J}_r\left(\boldsymbol{x}_1\right)^{-1} \boldsymbol{x}_2) & \boldsymbol{x}_2 \text { small }\end{cases}, \\
\mathrm{Exp} \left(\boldsymbol{x}_1 + \boldsymbol{x}_2\right) &\approx \begin{cases}\mathrm{Exp} (\boldsymbol{J}_{l}\left(\boldsymbol{x}_2\right) \boldsymbol{x}_1)\mathrm{Exp}(\boldsymbol{x}_2) & \boldsymbol{x}_1 \text { small } \\ \mathrm{Exp} (\boldsymbol{x}_1)\mathrm{Exp}(\boldsymbol{J}_r\left(\boldsymbol{x}_1\right) \boldsymbol{x}_2) & \boldsymbol{x}_2 \text { small }\end{cases},
\end{aligned}
\end{equation}
where $\bm{J}_l \in \mathbb{R}^{\dim \mathfrak{g} \times \dim \mathfrak{g}}$ and $\bm{J}_r \in \mathbb{R}^{\dim \mathfrak{g} \times \dim \mathfrak{g}}$ denote the left and right Jacobians of the Lie group, respectively.  

\subsection{Uncertainties on Lie Groups}
\label{sec:pre_b}
\begin{figure}[!t]
\centering
\includegraphics[width=0.36\textwidth]{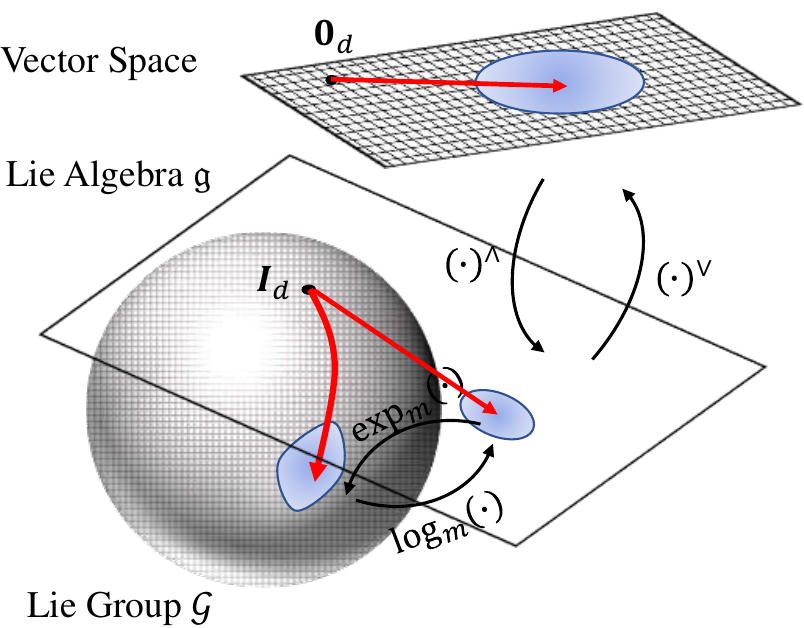}
\caption{Relationship among a matrix Lie group $\mathcal{G}$, its Lie algebra $\mathfrak{g}$, and the associated vector space $\mathbb{R}^{\dim \mathcal{G}}$. The color-graded ellipse depicts the uncertainty distribution.}
\label{fig:lie_group}
\end{figure}

State estimation requires characterizing the uncertainty of the state, which is typically assumed to be Gaussian variable \cite{barfoot2014associating,sjoberg2021lie, barfoot2024state}.
In Euclidean spaces, an \(n\)-dimensional Gaussian random variable \(\bm{x}\sim \mathcal{N}(\bar{\bm{x}}, \bm{\Sigma})\) can be written as $\bm{x} = \bar{\bm{x}} + \bm{\epsilon}$, where $\bm{\epsilon}$ satisfies $\bm{\epsilon} \sim \mathcal{N}(0,\bm{\Sigma})$ since Euclidean spaces are closed under addition. In contrast, Lie groups are closed under multiplication. Following the analogy with Euclidean spaces, a Gaussian random variable on a Lie group $\bm{X} \in \mathcal{G}$ can be expressed as
\begin{equation}
\label{eq.lie_group_gauss}
\bm{X}=\bm{\bar{X}}\mathrm{Exp}(\bm{\xi}), \quad \bm{\xi} \sim \mathcal{N}(0, \bm{\Sigma}),
\end{equation}
where $\bm{\bar{X}} \in \mathcal{G}$ is the noise-free mean, and \(\bm{\xi} \in \mathbb{R}^{\dim \mathcal{G}}\) is a Gaussian perturbation in the tangent space. This formulation is also referred to as the concentrated Gaussian distribution on Lie group \cite{barfoot2014associating}. Using the mapping in \eqref{eq.Exp_Log}, the group elements can be expressed as \(\bm{X} = \mathrm{Exp}(\bm{x})\) and \(\bm{\bar{X}} =\mathrm{Exp}(\bm{\bar{x}}) \). Applying the BCH formula \eqref{eq.BCH}, we obtain
\begin{equation}
\label{eq.eq.lie_vec_gauss}
\bm{x} \approx \bm{\bar{x}} + \bm{J}_r(\bm{\bar{x}})\bm{\xi},
\end{equation}
which implies \(\bm{x} \sim \mathcal{N}(\bm{\bar{x}},\bm{J}_r\bm{\Sigma}\bm{J}_r^{\top})\). Therefore, 
\eqref{eq.eq.lie_vec_gauss} characterizes the impact of a random Gaussian perturbation on the tangent space of a Lie group. These mappings, along with their uncertainty propagation, are illustrated in Fig.~\ref{fig:lie_group}.

\subsection{Typical Lie Groups}
The 3D orientation of a rigid body forms the special orthogonal group $\mathrm{SO}(3)$, defined as
\begin{equation}
\nonumber
\mathrm{SO}(3) := \left\{\bm{R}\in\mathbb{R}^{3\times3}\mid\bm{R}\bm{R}^\top=\mathrm{\bf{I}}_{3\times3}, \ \det\bm{R}=1\right\},
\end{equation}
which can be regarded as the set of all 3D rotation matrices. The Lie algebra \(\mathfrak{so}(3)\) corresponding to \(\mathrm{SO}(3)\) is
\begin{equation}
\nonumber
\begin{aligned}
\mathfrak{so}(3) := \left\{\boldsymbol{\phi}^{\wedge}\in\mathbb{R}^{3\times3} \mid \boldsymbol{\phi} \in \mathbb{R}^3 \right\}, 
\end{aligned}
\end{equation}
where the \textit{wedge} operator for \(\mathrm{SO}(3)\) maps a vector to a skew-symmetric matrix as
\begin{equation}
\nonumber
\\
\boldsymbol{\phi}^{\wedge}=\begin{bmatrix}0&-\phi_3&\phi_2\\\phi_3&0&-\phi_1\\-\phi_2&\phi_1&0\end{bmatrix}.
\end{equation}

Beyond orientation, state estimation in robotics often requires representing additional quantities such as position, velocity, and possibly landmarks in the environment \cite{zhang2017convergence}.  
When combined with attitude, these variables form another matrix Lie group, denoted as the special Euclidean group $\mathrm{SE}_m(3)$:
\begin{equation}
\nonumber
\begin{aligned}
\mathrm{SE}_m(3) := \big\{&\bm{X} = \begin{bmatrix}
\bm{R} & \bm{p}_1 & \ldots & \bm{p}_m \\
\bm{0}_{1\times 3} & 1 & \ldots & 0 \\
\vdots & \vdots & \ddots & \vdots \\
\bm{0}_{1\times 3} & 0 & \ldots & 1
\end{bmatrix} \in \mathbb{R}^{(3+m) \times (3+m)} \mid \\
&\bm{R} \in \mathrm{SO}(3),\quad \bm{p}_1, \ldots \bm{p}_m \in \mathbb{R}^3\big\},    
\end{aligned}
\end{equation}
where \(m \in \mathbb{N}^+\) specifies the number of additional translational components. Its Lie algebra \(\mathfrak{se}_m(3)\) is given by
\begin{equation}
\nonumber
\begin{aligned}
\mathfrak{se}_m(3) &:= \left\{\boldsymbol{\xi}^{\wedge}\in\mathbb{R}^{(3+m)\times(3+m)}\mid \boldsymbol{\xi}\in \mathbb{R}^{(3+m)}\right\}.
\end{aligned}
\end{equation}
The wedge operator for \(\mathrm{SE}_m(3)\) takes the form as
\begin{equation}
\nonumber
\begin{aligned}
\boldsymbol{\xi}^{\wedge}&=\begin{bmatrix}\bm{\phi}^{\wedge}&\bm{\rho}_1 & \ldots & \bm{\rho}_m\\\bm{0}_{m \times 3}& \bm{0}_{m \times 1} & \ldots & \bm{0}_{m \times 1}\end{bmatrix}, \\
\bm{\phi}^{\wedge} &= \log(\bm{R})\in \mathfrak{so}(3),\  \bm{\rho}_{i} = \bm{J}_l(\bm{\phi})^{-1}\bm{p}_i, i=1,\ldots, m.
\end{aligned}
\end{equation}
In addition, the adjoint matrix of $\mathrm{SE}_m(3)$ is also a commonly used tool in state estimation, which is expressed as
\begin{equation}
\nonumber
\mathrm{Ad}_{\mathbf{X}}=\begin{bmatrix}
\bm{R} & \bm{0}_{3\times3} & \cdots & \bm{0}_{3\times3} \\
\bm{p}_1^{\wedge} \bm{R} & \bm{R} & \cdots & \bm{0}_{3\times3} \\
\vdots & \vdots & \ddots & \vdots \\
\bm{p}_m^{\wedge} \bm{R} & \bm{0}_{3\times3} & \cdots & \bm{R}
\end{bmatrix} \in \mathbb{R}^{(3+3m)\times(3+3m)}.
\end{equation}

\section{System Modeling on Lie Groups}
\label{sec:modeling}
This section presents a modeling framework for robotic systems whose states evolve on Lie group manifolds. To ground our theoretical development in a practical and challenging application, we will use a legged robot state estimation task for illustration. 
This specific case was chosen for its challenging nature posed by discontinuous foot-ground contacts and potential measurement outliers from faulty contact detection \cite{bloesch2013state, hartley2020contact, gao2022invariant, yang2023multi}. Specifically, this section begins by defining the key state variables on the Lie group, followed by the derivation of the state propagation model and observation models, which together form the foundation for the proposed filter design.

\subsection{State Definition}
\label{sec_sys:a}
For a legged robot, the core states of interest are its body orientation \(\bm{R}_t \in \mathrm{SO}(3)\), velocity \(\bm{v}_t \in \mathbb{R}^3\) and position \(\bm{p}_t \in \mathbb{R}^3\) in the world frame. In addition, to utilize forward kinematics in constructing the state estimator, the positions of the contact points between robot's foot and ground \(\{\bm{s}_{i,t}\}_{i=1}^L\) are often incorporated into the state variables, where $L$ is the number of legs. Since the propagation and observation models for each contact point are identical and independent, we represent a generic contact point by a single variable \(\bm{s}_t \in \mathbb{R}^3\) without loss of generality. Collecting these components yields the full robot state, which admits a compact representation on a special Euclidean group:
\begin{equation}
\nonumber
\bm{X}_t = \begin{bmatrix}
\bm{R}_t & \bm{v}_t & \bm{p}_t & \bm{s}_t \\
\bm{0}_{3\times3} & \bm{0}_{3\times1} & \bm{0}_{3\times1} & \bm{0}_{3\times1} 
\end{bmatrix} \in \mathrm{SE}_3(3).
\end{equation}
This formulation unifies the robot’s orientation, velocity, position, and contact information within a single  group structure. 

\subsection{State Propagation Model}
\label{sec:sys_b}
The state propagation model describes how the robot’s state evolves over time, driven by control inputs from an onboard IMU. The control inputs, denoted as \(\bm{u}_t \in\mathbb{R}^6\), consist of the noisy body-frame linear acceleration $\tilde{\bm{a}}_t \in \mathbb{R}^{3}$ and angular velocity $\tilde{\bm{\omega}}_t \in \mathbb{R}^{3}$.  
They are corrupted by zero-mean white Gaussian noise \(\bm{n}^a_t \sim \mathcal{N}(\bm{0},\bm{\Sigma}_a) \) and \(\bm{n}^\omega_t \sim \mathcal{N}(\bm{0},\bm{\Sigma}_\omega)\), as:
\begin{equation}
\nonumber
    \tilde{\bm{a}}_t = \bm{a}_t + \bm{n}^a_t, \quad  \tilde{\bm{\omega}}_t = \bm{\omega}_t + \bm{n}^\omega_t ,
\end{equation}
From standard IMU kinematics \cite{barfoot2024state, hartley2020contact}, the continuous-time evolution of the pose and velocity is 
\begin{equation}
\label{eq.imu_dynamics}
\begin{aligned}
\frac{\mathrm{d}}{\mathrm{d}t}{\bm{R}_t} &= \bm{R}_t(\tilde{\bm{\omega}}_t -\bm{n}^{\omega})^{\wedge},\\ \frac{\mathrm{d}}{\mathrm{d}t}{\bm{p}_t} &= \bm{v}_t, \\
\frac{\mathrm{d}}{\mathrm{d}t}{\bm{v}_t} &= \bm{R}_t(\tilde{\bm{a}}_t -\bm{n}^{a}) + \bm{g},
\end{aligned}
\end{equation}
where \(\boldsymbol{g} \in \mathbb{R}^3\) is the gravitational acceleration. When a foot is in contact, its world-frame position is modeled as a small random walk to account for slippage: \(\frac{\mathrm{d}}{\mathrm{d}t} \bm{s}_t = \bm{R}_t\bm{n}^s_t\), where $\bm{n}^s \sim \mathcal{N}(0, \bm{\Sigma}_s)$ is also white Gaussian noise \cite{bloesch2013state,hartley2020contact, gao2022invariant}. 

The continuous propagation model can then be compactly expressed on the Lie group as
\begin{equation}
\label{eq.lie_process}
\begin{aligned}
\frac{\mathrm{d}}{\mathrm{d}t}\bm{X}_t &= \bm{f}_{\bm{u}_t}(\bm{X}_t) + \bm{X}_t\bm{n}_t^\wedge \\
&= \begin{bmatrix}
\bm{R}_t \tilde{\bm{\omega}}_t^\wedge  & \bm{R}_t \tilde{\bm{a}}_t + \bm{g} & \bm{v}_t & \bm{0}_{3 \times 1} \\
\bm{0}_{3\times3} & \bm{0}_{3\times1} & \bm{0}_{3\times1} & \bm{0}_{3\times1} 
\end{bmatrix} + \bm{X}_t\bm{n}_t^\wedge, 
\end{aligned}
\end{equation}
where \(\bm{f}_{\bm{u}_t}: \mathrm{SE}_3(3)  \rightarrow \mathrm{SE}_3(3)\) is the deterministic dynamics function, and \(\bm{n}_t = \left[(\bm{n}_t^{\omega})^\top, (\bm{n}_t^{a})^\top, \bm{0}_{3\times 1},(\bm{n}_t^{s})^\top\right]^\top \sim \mathcal{N}(0, \bm{Q}_t)\) represents the process noise with covariance \(\bm{Q}_t = \mathrm{diag}(\bm{\Sigma}_\omega, \bm{\Sigma}_a,\bm{0}_{3\times 3}, \bm{\Sigma}_s)\). The operator \(\mathrm{diag}(\cdot)\) constructs a block-diagonal matrix from its matrix-valued arguments. Finally, it is worth noting that the propagation model in \eqref{eq.lie_process} applies universally to all rigid-body robotic systems equipped with an IMU.


\subsection{Observation Model}
The observation model defines how the system state is observed through sensors, by identifying quantities that can be both computed from the state and from sensor measurements. For legged robots, joint encoders measure the leg joint angles, denoted by $\tilde{\bm{\theta}}_t \in \mathbb{R}^3$ for one leg, which are corrupted by Gaussian noise $\bm{n}^{\theta} \sim \mathcal{N}(\bm{0}, \bm{\Sigma}_\theta)$.  Substituting the noisy joint angles into the forward kinematics function \(\bm{fk}: \mathbb{R}^3 \rightarrow \mathbb{R}^3\), yields the foot contact point position relative to the robot body:
\begin{equation}
\nonumber
\begin{aligned}
\bm{r}_{\text{BF},t} &= \bm{fk}(\tilde{\bm{\theta}}_t - \bm{n}^{\theta}) \approx \bm{fk}(\tilde{\bm{\theta}}_t) - \bm{J}_{\text{fk}}(\tilde{\bm{\theta}}_t)\bm{n}^{\theta},
\end{aligned}
\end{equation}
where \(\bm{J}_{\text{fk}}(\tilde{\bm{\theta}}_t) \in \mathbb{R}^{3\times 3}\) is the kinematics Jacobian matrix. The explicit forms of \(\bm{fk}(\cdot)\) and \(\bm{J}_{\text{fk}}(\cdot)\) are given in Appendix~\ref{sec:appen}.  Meanwhile, the same relative position can also be expressed directly from the state variables, as
\begin{equation}
\nonumber
\bm{r}_{\text{BF},t} = \bm{R}_t^\top(\bm{s}_t - \bm{p}_t).
\end{equation}
By equating these two equivalent representations, the observation model is derived as
\begin{equation}
\label{eq.fk_meas}
\bm{fk}(\tilde{\bm{\theta}}_t) =\bm{R}_t^\top(\bm{s}_t - \bm{p}_t)+ \bm{J}_{\text{fk}}(\tilde{\bm{\theta}}_t)\bm{n}^{\theta}.
\end{equation}
This equation also corresponds to the leg odometry, which provides a relative position constraint for the robot by utilizing the contact point position and leg forward kinematics \cite{bloesch2013state, yang2023multi}. Given that the state evolves on a Lie group, the observation model can be expressed in a group-consistent form as
\begin{equation}
\label{eq.lie_meas}
\begin{aligned}
\underbrace{\begin{bmatrix}
\bm{fk}(\tilde{\bm{\theta}}_t) \\
0 \\
1 \\
-1 \\
\end{bmatrix}}_{\bm{y}_t} = \underbrace{\bm{X}_t^{-1}\underbrace{\begin{bmatrix}
  \bm{0}_{3\times1} \\
  0 \\
  1 \\
  -1 \\
\end{bmatrix}}_{\bm{b}}}_{\bm{h}(\bm{X}_t)} + \underbrace{\begin{bmatrix}
  \bm{J}_{\text{fk}}(\tilde{\bm{\theta}}_t)\bm{n}^{\theta} \\
  0 \\
  0 \\
  0 \\
\end{bmatrix}}_{\bm{\eta}_t}.   
\end{aligned}
\end{equation}
Here, \(\bm{h}: \mathrm{SE}_3(3) \rightarrow \mathbb{R}^6\) denotes the measurement function, \(\bm{y}_t \in \mathbb{R}^6\) is the noisy measurement,  and \(\bm{\eta}_t \sim \mathcal{N}(0, \bm{\Gamma}_t)\) represents the measurement noise, whose covariance is given by \(\bm{\Gamma}_t=\mathrm{diag}(\bm{J}_{\text{fk}}(\tilde{\bm{\theta}}_t)\bm{\Sigma}_\theta \bm{J}_{\text{fk}}^\top(\tilde{\bm{\theta}}_t), \bm{0}_{3\times 3})\).

The observation model in \eqref{eq.lie_meas} is highly general, extending beyond leg odometry to any sensor modality that provides relative position constraints, such as visual, LiDAR, or wheel odometry. For example, in LiDAR odometry, a stable environmental feature identified via scan-matching serves the same role as the stationary foot contact point, fitting directly into the model's mathematical structure \cite{xu2021fast}. Finally, observation models with such a mathematical structure are generally referred to as invariant observation models \cite{barrau2016invariant,hartley2020contact}.

\section{Natural gradient Gaussian Approximation on Lie Group Filter}
In the previous section, we established a state space model for robot state estimation on the Lie group manifold $\mathcal{G}=\mathrm{SE}_3(3)$. Building on this foundation, this section introduces our proposed filtering algorithm, which unfolds in three stages. First, in Sec.~\ref{sec:method_a}, we reframe the manifold filtering task as a Gaussian parameter optimization problem. Following this, we detail the two key steps of our solution framework in Sec.~\ref{sec:method_b} and Sec.~\ref{sec:method_c}.

\subsection{Optimization Objective Formulation}
\label{sec:method_a}

The objective of state estimation is to recover the system state $\bm{X}_t$ from noisy measurements $\bm{y}_{t}$. Typically, filtering algorithms achieve this by a two-step procedure: (i) the \textit{prediction} step computes a prior estimate \(\hat{\bm{X}}_{t|t-1}\), using the state propagation model \eqref{eq.lie_process} together with the control input \(\bm{u}_{t-1}\), and (ii) the \textit{update} step refines this estimate into the final posterior estimate \(\hat{\bm{X}}_{t|t}\) by incorporating the observation model \eqref{eq.lie_meas} and the noisy observation \(\bm{y}_t\). 

On Lie group manifolds, the prior estimate \(\hat{\bm{X}}_{t|t-1}\) can be readily obtained by applying a forward Euler integration step to the noise-free system dynamics, \(\frac{\mathrm{d}}{\mathrm{d}t}\bm{X}_t = \bm{f}_{\bm{u}_t}(\bm{X}_t)\), from the  posterior estimate \(\hat{\bm{X}}_{t-1|t-1}\) in the last time step \cite{barrau2016invariant, hartley2020contact, gao2022invariant}.
In contrast, solving for the posterior is more challenging because the Lie group operation is not closed under addition, preventing the direct use of additive update rules from classical Euclidean filters. 
Nevertheless, the same intuition holds: the update step essentially involves estimating a state increment relative to the prior, based on the measurements \cite{markley2003attitude, barrau2016invariant}. 
In the case of Lie group, since the increment is typically small, it can be fully characterized within the tangent space. 
Therefore, we define  \(\hat{\bm{\xi}}_{t|t} \in \mathbb{R}^{d}\) as the posterior estimate of increment, where \(d=\dim \mathcal{G}\) denotes the dimension of the Lie group’s tangent space. By exploiting the closure property of Lie groups under multiplication and the mapping in \eqref{eq.Exp_Log}, the posterior update takes the compact form:
\begin{equation}
\label{eq.state_increment}
\hat{\bm{X}}_{t|t} = \hat{\bm{X}}_{t|t-1}\mathrm{Exp}(\hat{\bm{\xi}}_{t|t}).
\end{equation}
This formulation transforms the problem of estimating the full state posterior \(\hat{\bm{X}}_{t|t}\) into the more tractable problem of estimating the increment posterior \(\hat{\bm{\xi}}_{t|t}\) \cite{barrau2016invariant, brossard2020code, goffin2024iterated}.


According to Bayesian theory, obtaining the posterior estimate of the increment requires computing its posterior distribution \cite{knoblauch2022optimization, goffin2024iterated}, which can be expressed as
\begin{equation}
\label{eq.bayes}
p(\bm{\xi}_t|\bm{y}_{1:t},\bm{u}_{0:t-1}) = \frac{p(\bm{\xi}_t|\bm{y}_{1:t-1}, \bm{u}_{0:t-1})p_{\hat{\bm{X}}_{t|t-1}}(\bm{y}_t|\bm{\xi}_t)}{\int{p(\bm{\xi}_t|\bm{y}_{1:t-1}, \bm{u}_{0:t-1})p_{\hat{\bm{X}}_{t|t-1}}(\bm{y}_t|\bm{\xi}_t)}\mathrm{d}\bm{\xi}_t}.
\end{equation}
Here, \(p(\bm{\xi}_t|\bm{y}_{1:t-1}, \bm{u}_{0:t-1})\) denotes the prior distribution of the increment, and \(p_{\hat{\bm{X}}_{t|t-1}}(\bm{y}_t|\bm{\xi}_t)\) represents the measurement likelihood. Based on the update rule in \eqref{eq.state_increment} and the observation model in \eqref{eq.lie_meas}, the likelihood takes a Gaussian form:
\begin{equation}
\label{eq.likelihood}
p_{\hat{\bm{X}}_{t|t-1}}(\bm{y}_t|\bm{\xi}_t) = \mathcal{N}(\bm{y}_t;\mathrm{Exp}(-\bm{\xi}_t)\hat{\bm{X}}_{t|t-1}^{-1}\bm{b}, \bm{\Gamma}_t).
\end{equation}
Since the likelihood is nonlinear with respect to \(\bm{\xi}_t\), the integral in the denominator of \eqref{eq.bayes} becomes intractable, making direct posterior computation infeasible \cite{knoblauch2022optimization}. Existing methods typically sidestep this issue by locally approximating the measurement likelihood via Taylor series expansions \cite{bourmaud2013discrete, bourmaud2015continuous,barrau2016invariant, goffin2024iterated} or UT \cite{brossard2017unscented, brossard2020code, sjoberg2021lie,jin2024nonlinear}. However, these approximations inevitably introduce errors due to the linearization or sampling assumptions involved \cite{cao2024nonlinear}.

Instead of relying on local approximations of the measurement likelihood, we propose to optimize the posterior distribution directly using the optimization perspective of Bayesian inference \cite{knoblauch2022optimization,cao2024nonlinear}. Under this framework, the posterior distribution is characterized as the solution to an optimization problem:
\begin{equation}
\label{eq.variational}
\begin{aligned}
p(\bm{\xi}_t|\bm{y}_{1:t},\bm{u}_{0:t-1}) 
=&\mathop{\arg\min}\limits_{q(\bm{\xi}_t)} \big\{\mathbb{E}_{q(\bm{\xi}_t)}\left[-\log{p_{\hat{\bm{X}}_{t|t-1}}(\bm{y_t}|\bm{\xi}_t)}\right] \\
&+ \mathcal{D}_{\text{KL}}\left[q(\bm{\xi}_t)\|p(\bm{\xi}_t|\bm{y}_{1:t-1},\bm{u}_{0:t-1})\right] \big\},
\end{aligned}
\end{equation}
where \(q(\bm{\xi}_t)\) is the variational distribution, \(\mathcal{D}_{\text{KL}}[\cdot\|\cdot]\) is Kullback-Leibler(KL) divergence between two distributions.  
This formulation essentially balances the prior information with the current measurement information \cite{knoblauch2022optimization}. However, directly solving \eqref{eq.variational} requires finding an optimal variational distribution over the entire probability space, which is generally intractable. A tractable alternative is to restrict the variational distribution to a parameterizable family of distributions. Drawing from the properties discussed in Sec.~\ref{sec:pre_b} and building on established work \cite{bourmaud2013discrete, barrau2016invariant, sjoberg2021lie}, the Gaussian distribution family is the most natural choice. Accordingly, we define the variational distribution as \(q(\bm{\xi}_t)=\mathcal{N}(\bm{\xi}_t;\hat{\bm{\xi}}_t, \bm{P}_t)\) and approximate both the prior and posterior distributions by Gaussians:
\begin{equation}
\nonumber
\begin{aligned}
p(\bm{\xi}_{t}|\bm{y}_{1:t-1},\bm{u}_{0:t-1}) &\approx \mathcal{N}(\bm{\xi}_{t};\hat{\bm{\xi}}_{t|t-1}, \bm{P}_{t|t-1}), \\ p(\bm{\xi}_t|\bm{y}_{1:t},\bm{u}_{0:t-1}) &\approx \mathcal{N}(\bm{\xi}_{t};\hat{\bm{\xi}}_{t|t}, \bm{P}_{t|t})
\end{aligned}    
\end{equation}
Here, $\hat{\bm{\xi}}_{t|t-1} \in \mathbb{R}^d$ and $\hat{\bm{\xi}}_{t|t} \in \mathbb{R}^d$ are the prior and posterior estimates of increment, $\bm{P}_{t|t-1} \in \mathbb{R}^{d\times d}$ and $\bm{P}_{t|t}\in \mathbb{R}^{d\times d}$ are the prior and posterior covariance matrices, respectively.  

By substituting these Gaussian distributions into \eqref{eq.variational}, the infinite-dimensional variational inference problem reduces to a tractable, finite-dimensional optimization problem over the Gaussian parameters:
\begin{equation}
\label{eq.param_optimization}
\begin{aligned}
\hat{\bm{\xi}}_{t|t}, \bm{P}_{t|t} &=  \mathop{\arg\min}\limits_{\hat{\bm{\xi}}_{t}, \bm{P}_{t}}\ J(\hat{\bm{\xi}}_{t}, \bm{P}_{t}),\\
J(\hat{\bm{\xi}}_{t}, \bm{P}_{t}) &=\mathbb{E}_{\mathcal{N}(\bm{\xi}_t;\hat{\bm{\xi}}_{t}, \bm{P}_t)} \left[-\log{p_{\hat{\bm{X}}_{t|t-1}}(\bm{y_t}|\bm{\xi}_t)}\right]  \\
&+\frac{1}{2} \left(\hat{\bm{\xi}}_{t|t-1}-\hat{\bm{\xi}}_{t} \right)^{\top} \bm{P}_{t|t-1}^{-1}\left(\hat{\bm{\xi}}_{t|t-1}-\hat{\bm{\xi}}_{t} \right)
\\
&+ \frac{1}{2} \mathrm{Tr}\left(\bm{P}_{t|t-1}^{-1} \bm{P}_t \right)
-\frac{1}{2} \log \frac{\left|\bm{P}_t\right|}{\left|\bm{P}_{t|t-1}\right|}-\frac{1}{2}d, \\
\end{aligned}
\end{equation}
Consequently, the estimation procedure consists of two steps: (a) computing the Gaussian parameters of the increment prior \(\hat{\bm{\xi}}_{t|t-1}\) and \(\bm{P}_{t|t-1}\); (b) substituting them into \eqref{eq.param_optimization} to solve for the Gaussian parameters of the increment posterior \(\hat{\bm{\xi}}_{t|t}\) and \(\bm{P}_{t|t}\). To avoid confusion with the prediction and update steps of the full state, we refer to these two stages as \textit{increment prior computation} and \textit{increment posterior optimization}, respectively. Together, they constitute the update step of the full-state filter via \eqref{eq.state_increment}. The following sections detail the solution procedure for these two stages.


\subsection{Increment Prior Computation}
\label{sec:method_b}
Analogous to the computation of the full-state prior, deriving the increment prior involves propagating the posterior from the previous time step through the increment dynamics model.
Although the underlying IMU-driven system dynamics \eqref{eq.lie_process} is highly nonlinear and defined on the Lie group, it can be simplified into a much simpler dynamics model for the increment by exploiting the group-affine property \cite{barrau2016invariant, hartley2020contact, gao2022invariant}.

A system is said to be \emph{group-affine} \cite{ barrau2016invariant,hartley2020contact} if, for all \(\bm{X}_1,\bm{X}_2 \in \mathcal{G}\), the dynamics function \(f_{\bm{u}_t}(\cdot)\) satisfies
\begin{equation}
\nonumber
\begin{aligned}
f_{\bm{u}_t}(\bm{X}_1\bm{X}_2) &= f_{\bm{u}_t}(\bm{X}_1)\bm{X}_2+\bm{X}_1f_{\bm{u}_t}(\bm{X}_2)-\bm{X}_1f_{\bm{u}_t}(\bm{\mathrm{I}}_d)\bm{X}_2,
\end{aligned}
\end{equation}
where \(\bm{I}_d \in \mathcal{G}\) is the group identity. The IMU-driven system \eqref{eq.lie_process} can be verified to satisfy this property \cite{hartley2020contact, gao2022invariant}.
For such systems, it can be shown that the propagation model of the increment is approximately linear \cite{barrau2016invariant, hartley2020contact, gao2022invariant}:
\begin{equation}
\label{eq.increment_propagation}
\frac{\mathrm{d}}{\mathrm{d}t}\bm{\xi}_t = \bm{F} \bm{\xi}_t + \mathrm{Ad}_{\hat{\bm{X}}_{t|t-1}}\bm{n}_t,
\end{equation}
where \(\mathrm{Ad}_{\hat{\bm{X}}_{t|t-1}}\) is the adjoint matrix of \(\hat{\bm{X}}_{t|t-1}\) and \(\bm{F} \in \mathbb{R}^{d \times d}\) is a time-invariant matrix, given by
\begin{equation}
\nonumber
\bm{F} = \begin{bmatrix}
\bm{0} & \bm{0} & \bm{0} & \bm{0} \\
(\bm{g})^{\wedge} & \bm{0} & \bm{0} & \bm{0} \\
\bm{0} & \bm{\mathrm{I}} & \bm{0} & \bm{0} \\
\bm{0} & \bm{0} & \bm{0} & \bm{0} \\
\end{bmatrix}. 
\end{equation}
Discretizing \eqref{eq.increment_propagation} yields
\begin{equation}
\nonumber
\bm{\xi}_{t+1} = \bm{A}\bm{\xi}_t + \bm{B}_t\bm{n}_t,
\end{equation}
where \(\bm{A} = \bm{I}+\bm{F}\Delta t\) and \(\bm{B}_t = (\bm{I}+\bm{F}\Delta t) \mathrm{Ad}_{\hat{\bm{X}}_{t|t-1}}\).
Thus, the prior Gaussian mean and covariance can be propagated directly using standard Kalman filter formula as
\begin{equation}
\label{eq.predict_invariant}
\begin{aligned}
 \hat{\bm{\xi}}_{t|t-1} &= \bm{A}\hat{\bm{\xi}}_{t-1|t-1}, \\
\bm{P}_{t|t-1} &= \bm{A}\bm{P}_{t-1|t-1}\bm{A}^\top + \bm{B}_t\bm{Q}_t\bm{B}_t^\top.
\end{aligned}
\end{equation}

\begin{figure}[!t]
\centering
\includegraphics[width=0.45\textwidth]{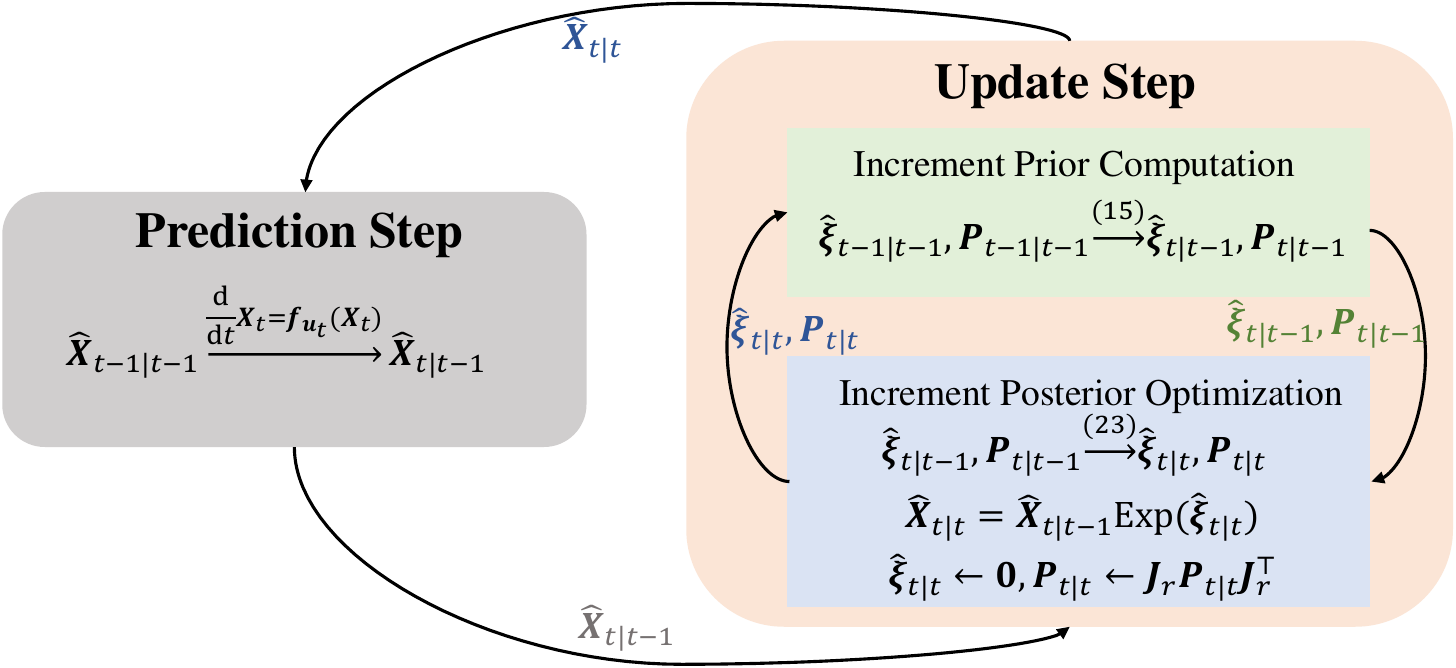}
\caption{The overall estimation procedure of our NANO-L filter at time step \(t\).}
\label{fig:architecture}
\end{figure}

\subsection{Increment
Posterior Optimization}
\label{sec:method_c}
With the prior distribution established, this stage reduces to solving the optimization problem in \eqref{eq.param_optimization}. Inspired by previous works on Gaussian variational inference \cite{wuunderstanding, cao2024nonlinear, barfoot2024state}, we use natural gradient descent (NGD) to optimize the cost function \(J(\hat{\bm{\xi}}_t, \bm{P}_t)\) in \eqref{eq.param_optimization}. The NGD leverages information geometry to compute the steepest descent direction in the Gaussian parameters space, leading to more efficient updates compared to standard gradient descent \cite{amari1998natural, martens2020new}.

For simplicity, we first denote the negative log-likelihood term in \eqref{eq.param_optimization} as
\(\ell(\bm{\xi}_t, \bm{y}_t)\). Substituting \eqref{eq.likelihood} into this expression yields
\begin{equation}
\label{eq.log_likelihood}
\begin{aligned}
\ell(\bm{\xi}_t, \bm{y_t})
=&-\log{p_{\hat{\bm{X}}_{t|t-1}}(\bm{y}_t|\bm{\xi}_t)} \\
=& \frac{1}{2}(\bm{y}_t - \mathrm{Exp}(-\bm{\xi}_t)\hat{\bm{X}}_{t|t-1}^{-1}\bm{b})^{\top} \bm{\Gamma}_t^{-1}\cdot \\
&(\bm{y}_t - \mathrm{Exp}(-\bm{\xi}_t)\hat{\bm{X}}_{t|t-1}^{-1}\bm{b}) + \mathrm{const}.
\end{aligned}
\end{equation}
Furthermore, we stack the variational Gaussian parameters into a single column vector $\bm{\psi}$ and compute the gradient with respect to it:
\begin{equation}
\nonumber
\begin{aligned}
\bm{\psi} = \begin{bmatrix}
\hat{\bm{\xi}}_t \\ \rvec(\bm{P}_t^{-1})
\end{bmatrix}, \ \nabla_{\bm{\psi}} J = \begin{bmatrix}
\frac{\partial}{\partial \hat{\bm{\xi}}_t} J(\hat{\bm{\xi}}_t, \bm{P}_t)  \\ \rvec\left(\frac{\partial}{\partial (\bm{P}_t)^{-1}} J(\hat{\bm{\xi}}_t, \bm{P}_t)\right)
\end{bmatrix}.
\end{aligned}
\end{equation}
Here, the partial derivatives of \(J(\hat{\bm{\xi}}_t, \bm{P}_t)\) with respect to \(\hat{\bm{\xi}}_t\) and \(\bm{P}_t\) are given by
\begin{equation}
\label{eq.derivate_gaussian}
\begin{aligned}
&\frac{\partial J(\hat{\bm{\xi}}_t, \bm{P}_t)}{\partial \hat{\bm{\xi}}_t} 
\\
=& \mathbb{E}_{\mathcal{N}(\bm{\xi}_t; \hat{\bm{\xi}}_t, \bm{P}_t)} \left\{ \frac{\partial \ell(\bm{\xi}_t, \bm{y}_t)}{\partial \bm{\xi}_t} \right\} + \bm{P}_{t|t-1}^{-1}(\hat{\bm{\xi}}_t - \hat{\bm{\xi}}_{t|t-1}),
\\
&\frac{\partial J(\hat{\bm{\xi}}_t, \bm{P}_t)}{\partial \bm{P}_t^{-1}} 
\\
=& \frac{1}{2}\bm{P}_t -\frac{1}{2} \bm{P}_t \big(\mathbb{E}_{\mathcal{N}(\bm{\xi}_t; \hat{\bm{\xi}}_t, \bm{P}_t)} \left\{ \frac{\partial^2 \ell(\bm{\xi}_t, \bm{y}_t)}{\partial \bm{\xi}_t^2} \right\} + \bm{P}^{-1}_{t|t-1}\big) \bm{P}_t,
\\
\end{aligned}
\end{equation}

The NGD update scheme can then be defined as
\begin{equation}
\label{eq.natural_gradient}
\bm{\psi}^{(i+1)} = \bm{\psi}^{(i)}  -\mathcal{F}^{-1}_{\bm{\psi}}\nabla_{\bm{\psi}} J,
\end{equation}
where \(\mathcal{F}^{-1}_{\bm{\psi}}\) denotes the inverse of the Fisher information matrix, specified as
\begin{equation}
\label{eq.fisher}
\begin{aligned}
\mathcal{F}^{-1}_{\bm{\psi}} = 
\begin{bmatrix}
\bm{P}_t & 0 \\
0 & 2 ((\bm{P}_t)^{-1} \otimes (\bm{P}_t)^{-1})
\end{bmatrix}, 
\end{aligned}   
\end{equation}
with $\otimes$ denoting the Kronecker product \cite{barfoot2024state, cao2024nonlinear}.
By substituting the Fisher information matrix and the partial derivatives into the update rule \eqref{eq.natural_gradient}, we obtain the iterative update equations:
\begin{equation}
\label{eq.iter}
\begin{aligned}
(\bm{P}_{t}^{-1})^{(i+1)} =& \bm{P}_{t|t-1}^{-1} + \mathbb{E}_{\mathcal{N}(\bm{\xi}_t;\hat{\bm{\xi}}_t^{(i)}, \bm{P}_t^{(i)})}\big\{\frac{\partial^2 \ell(\bm{\xi}_t, \bm{y}_t)}{\partial\bm{\xi}_t^\top\partial \bm{\xi}_t}\big\}  \\
\hat{\bm{\xi}}_t^{(i+1)} =& \hat{\bm{\xi}}_t^{(i)} - \bm{P}_{t}^{(i+1)}\mathbb{E}_{\mathcal{N}(\bm{\xi}_t;\hat{\bm{\xi}}_t^{(i)}, \bm{P}_t^{(i)})}\big\{\frac{\partial \ell(\bm{\xi}_t, \bm{y}_t)}{\partial \bm{\xi}_t}\big\} \\
&- \bm{P}_{t}^{(i+1)}\bm{P}_{t|t-1}^{-1}(\hat{\bm{\xi}}_t^{(i)}-\hat{\bm{\xi}}_{t|t-1}).
\end{aligned}
\end{equation} 
This iterative scheme drives the Gaussian parameters to stably converge \cite{barfoot2020exactly,cao2024nonlinear}. However, the expectations of the two derivatives of \(\ell\) (Hessian and Jacobian, respectively) generally have no closed-form solution and must be approximated using sampling-based techniques such as unscented transformation and cubature integration \cite{julier2004unscented,arasaratnam2009cubature}. In particular, computing the Hessian term incurs a computational complexity of \(\mathcal{O}(d^3)\), which grows rapidly with the state dimension. For practical systems with high-dimensional states, such as legged robots here, this computational burden becomes prohibitively expensive, making real-time deployment of the iterative update scheme infeasible.

To alleviate this computational burden, we observe that for the negative log-likelihood defined by the invariant observation model like \eqref{eq.log_likelihood}, its derivatives with respect to \(\bm{\xi}\) can be derived analytically by leveraging the perturbation model on Lie derivatives, yielding
\begin{equation}
\label{eq.derivative}
\begin{aligned}
\frac{\partial \ell(\bm{\xi}_t, \bm{y}_t)}{\partial \bm{\xi}_t} &= (\hat{\bm{X}}_{t|t-1}^{-1}\bm{b})^\odot\bm{\Gamma}_t^{-1}(\bm{y}_t - \mathrm{Exp}(-\bm{\xi}_t)\hat{\bm{X}}_{t|t-1}^{-1}\bm{b}), \\
\frac{\partial^2 \ell(\bm{\xi}_t, \bm{y}_t)}{\partial\bm{\xi}_t^\top\partial \bm{\xi}_t} &=  {(\hat{\bm{X}}_{t|t-1}^{-1}\bm{b})^\odot} \bm{\Gamma}_t^{-1} {(\hat{\bm{X}}_{t|t-1}^{-1}\bm{b})^\odot}^\top, 
\end{aligned}
\end{equation}
where the explicit form of \((\bm{X}^{-1}\bm{b})^\odot\) is given by 
\begin{equation}
(\bm{X}^{-1}\bm{b})^\odot = \begin{bmatrix}
\big(\bm{R}(\bm{p} - \bm{s})\big)^\wedge & \bm{0}_{3 \times 3} & \bm{\mathrm{I}}_{3\times 3} & -\bm{\mathrm{I}}_{3\times 3} \\
\bm{0}_{3 \times 3} & \bm{0}_{3 \times 3} & \bm{0}_{3 \times 3} & \bm{0}_{3 \times 3}.
\end{bmatrix}^\top
\end{equation}
The detailed derivation of these derivatives is provided in Appendix~\ref{sec:appen_b}.
Here, the Hessian of $\ell$ is an analytic expression that is independent of \(\bm{\xi}_t\). Therefore, the Hessian's expectation is simply the Hessian itself, which eliminates the need for costly sampling-based computation. Substituting these analytic derivatives into \eqref{eq.iter} yields the final NGD update:
\begin{equation}
\label{eq.final_iter}
\begin{aligned}
(\bm{P}_{t}^{-1})^{(i+1)} &= \bm{P}_{t|t-1}^{-1} + {(\hat{\bm{X}}_{t|t-1}^{-1}\bm{b})^\odot} \bm{\Gamma}_t^{-1} {(\hat{\bm{X}}_{t|t-1}^{-1}\bm{b})^\odot}^\top,   \\
\hat{\bm{\xi}}_t^{(i+1)} &= \hat{\bm{\xi}}_t^{(i)}  - \bm{P}_{t}\bm{P}_{t|t-1}^{-1}(\hat{\bm{\xi}}_t^{(i)}-\hat{\bm{\xi}}_{t|t-1})- \bm{P}_{t}(\hat{\bm{X}}_{t|t-1}^{-1}\bm{b})^\odot   \\
&\cdot \bm{\Gamma}_t^{-1}(\bm{y}_t -\mathbb{E}_{\mathcal{N}(\bm{\xi}_t;\hat{\bm{\xi}}_t^{(i)}, \bm{P}_t)}\{\mathrm{Exp}(-\bm{\xi}_t)\hat{\bm{X}}_{t|t-1}^{-1}\bm{b}\}).
\end{aligned}
\end{equation}
Unlike the general iterative form in \eqref{eq.iter}, the covariance update in \eqref{eq.final_iter} admits a closed-form solution and requires no iteration, while only the mean \(\hat{\bm{\xi}}_t\) is refined iteratively. This significantly reduces computational complexity and enables real-time applicability.

For this iterative scheme, the initial mean and covariance are set to the prior parameters, i.e., \(\hat{\bm{\xi}}_t^{(0)} = \hat{\bm{\xi}}_{t|t-1}\), \(\bm{P}_t^{(0)} = \bm{P}_{t|t-1}\). The iteration proceeds until the KL divergence between two successive Gaussian distributions falls below a predefined threshold \(\gamma\) as
\begin{equation}
\label{eq.terminate}
\mathcal{D}_{\mathrm{KL}}(\mathcal{N}(\hat{\bm{\xi}}^{(i)}_t, \bm{P}^{(i)}_t)\| \mathcal{N}(\hat{\bm{\xi}}^{(i+1)}_t, \bm{P}^{(i+1)}_t)) < \gamma.   
\end{equation}
Once this condition is satisfied, the algorithm is regarded as converged, and the posterior mean and covariance are obtained as
\(\hat{\bm{\xi}}_{t|t} = \hat{\bm{\xi}}_t^{(i+1)}, \quad \bm{P}_{t|t} = \bm{P}_{t}^{(i+1)}\).

Finally, the posterior Gaussian parameters of increment are lifted back onto the Lie group manifold using \eqref{eq.state_increment} to update the full-state:
\begin{equation}
\label{eq.final_equation}
\begin{aligned}
 &\hat{\bm{X}}_{t|t-1}\mathrm{Exp}(\hat{\bm{\xi}}_{t|t} + \bm{P}_{t|t}^{1/2}\bm{\epsilon})\\
 &= \underbrace{\hat{\bm{X}}_{t|t-1}\mathrm{Exp}(\hat{\bm{\xi}}_{t|t})}_{\hat{\bm{X}}_{t|t}}\mathrm{Exp}(\underbrace{\bm{J}_r(\hat{\bm{\xi}}_{t|t})\bm{P}_{t|t}^{1/2}\bm{\epsilon}}_{\bm{\xi}_{t|t}}),.
\end{aligned}
\end{equation}
Here, \(\hat{\bm{X}}_{t|t}\) denotes the full-state posterior estimate, \(\bm{\epsilon}\in \mathcal{N}(\bm{0}, \bm{I})\) is a standard Gaussian random variable. The posterior uncertainty is represented in the tangent space with covariance \(\bm{P}_{t|t} \leftarrow \bm{J}_r(\hat{\bm{\xi}}_{t|t})\bm{P}_{t|t}\bm{J}_r(\hat{\bm{\xi}}_{t|t})^\top\), which captures the confidence associated with the estimate. Finally, the increment is reset to the origin \(\hat{\bm{\xi}}_{t|t}=\bm{0}\), to prepare for the next prediction cycle. The complete procedure constitutes the NANO-L filter, summarized in Fig~\ref{fig:architecture} and Algorithm~\ref{alg.1}.

\begin{algorithm}
\caption{NANO-L filter}\label{alg.1}
\renewcommand{\algorithmicrequire}{\textbf{Input:}}
\renewcommand{\algorithmicensure}{\textbf{Output:}}
\begin{algorithmic}[1]

\REQUIRE Initial state $\hat {\boldsymbol{X}}_{0|0}$, initial covariance $\bm{P}_{0|0}$, hyperparameter $\gamma$

\ENSURE State estimates $\{\hat {\bm{X}}_{t|t}\}_{t=1}^{T}$
\FOR{each time step $t$}
    \STATE \textbf{Perform Prediction step:}
        \STATE Compute prior full-state estimate $\hat{\bm{X}}_{t|t-1}$ by integrating \(\frac{\mathrm{d}}{\mathrm{d}t}\bm{X}_t = \bm{f}_{\bm{u}_t}(\bm{X}_t)\)
    \STATE \textbf{Perform Update step:}
    \STATE Compute prior increment estimate \(\hat{\bm{\xi}}_{t|t-1}\) and its covariance \(\bm{P}_{t|t-1}\) using \eqref{eq.predict_invariant}
    \STATE Initialize the iterative Gaussian parameters: \(\hat{\bm{\xi}}_{t}^{(0)} \gets \hat{\bm{\xi}}_{t|t-1}\), \(\bm{P}_t^{(0)} \gets \bm{P}_{t|t-1}\)
    \STATE Perform natural gradient iterations of \eqref{eq.final_iter} until the stopping criterion \eqref{eq.terminate} is satisfied, yielding the posterior increment estimate \(\hat{\bm{\xi}}_{t|t}\) and its covariance \(\bm{P}_{t|t}\).
    \STATE Obtain the posterior full-state estimate \(\hat{\bm{X}}_{t|t}\) and its covariance using \eqref{eq.final_equation}.
    
    
\ENDFOR
\end{algorithmic}
\end{algorithm}

\section{Numerical Simulations}
To validate the performance of the proposed NANO-L filter in a controlled and repeatable setting, we first conduct a numerical simulation. The scenario involves a three-dimensional rigid-body robot equipped with an IMU, whose state is corrected by vision-based observations of static, known landmarks in the environment. 
This simulation can be regarded as making two simplifications to the modeling presented in Section~\ref{sec:modeling}. First, the state propagation model uses only the IMU kinematics \eqref{eq.imu_dynamics}, excluding the foot contact states. Second, the observation model no longer depends on a moving foot contact point \(\bm{s}_t\), but instead directly measures the relative position of three static landmarks (\(\bm{m}_1=[0, 2, 2]^\top\), \(\bm{m}_2=[-2, -2, -2]^\top\), \(\bm{m}_3=[2, -2, -2]^\top\)). The detailed model equations are shown in the Appendix~\ref{sec:appen_c}. A ground-truth trajectory was generated, and synthetic IMU and camera measurements were created by sampling the corresponding models and adding pre-defined Gaussian noise.

To mitigate the influence of randomness, we performed 100 Monte Carlo (MC) trials, with each trial simulating 30 seconds of motion at 100 Hz. We benchmark our NANO-L filter against a comprehensive set of baselines. These include classic Euclidean filters: EKF, UKF and NANO filter \cite{cao2024nonlinear}, as well as two commonly used filters designed for Lie group manifolds: InEKF \cite{barrau2016invariant, zhu2022design} and UKF-M \cite{brossard2020code}. The chosen evaluation metrics are the root mean square error (RMSE) for both orientation and position, following the formulation in \cite{brossard2020code}.

\begin{figure}[!t]
\centering
\subfloat{
    \includegraphics[width=0.4 \textwidth]{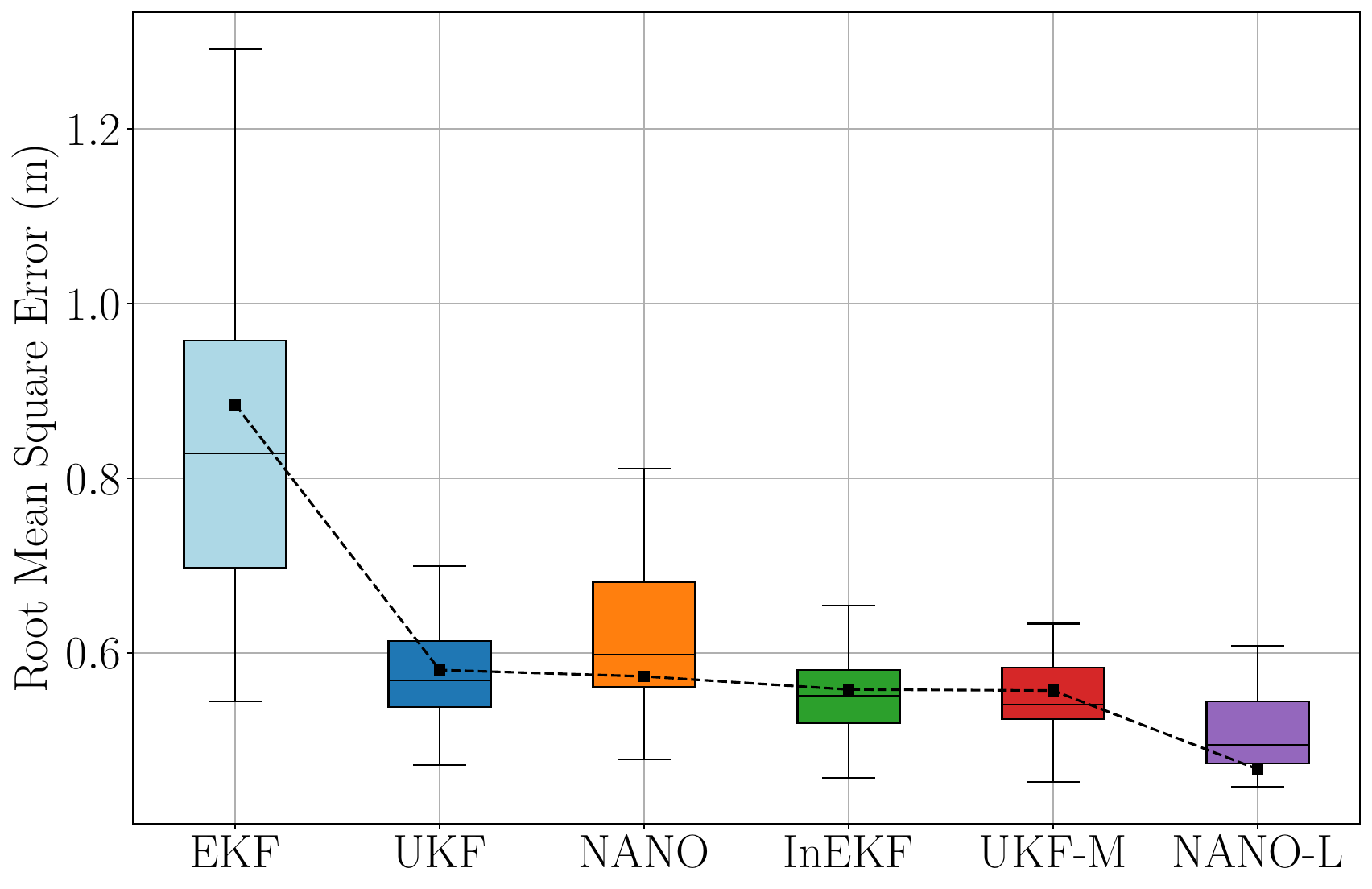}
    \label{fig:inertial_pos_rmse}
}
\\
\subfloat{
    \includegraphics[width=0.4\textwidth]{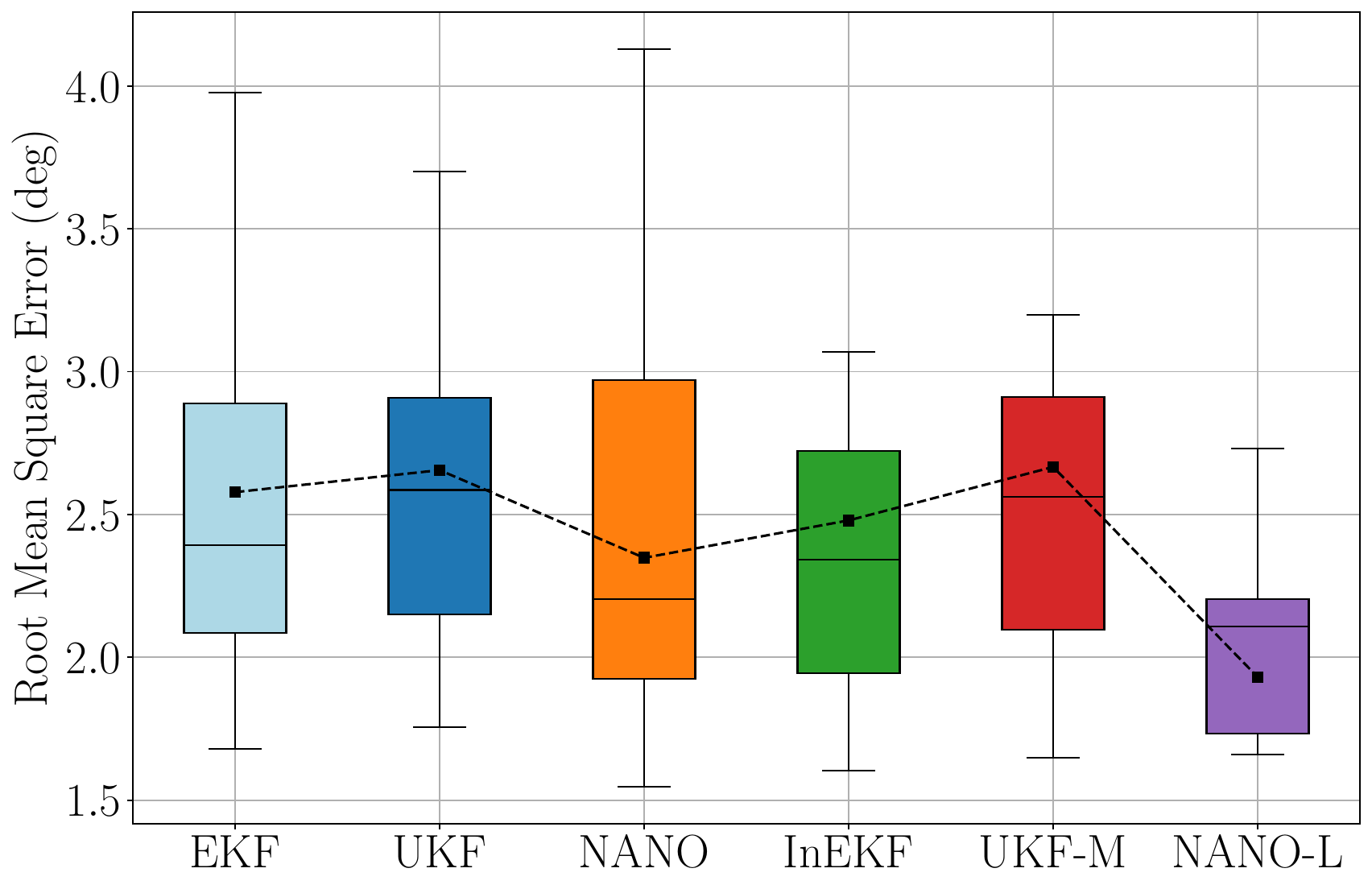}
    \label{fig:inertial_ori_rmse}
}
\caption{Box plots of position and orientation RMSE, aggregated over 100 Monte Carlo simulation trials. Our NANO-L filter achieves the lowest median error and smallest variance.}
\label{fig:inertial_rmse}
\end{figure}

\begin{figure}[!t]
\centering
\subfloat{
    \includegraphics[width=0.38\textwidth]{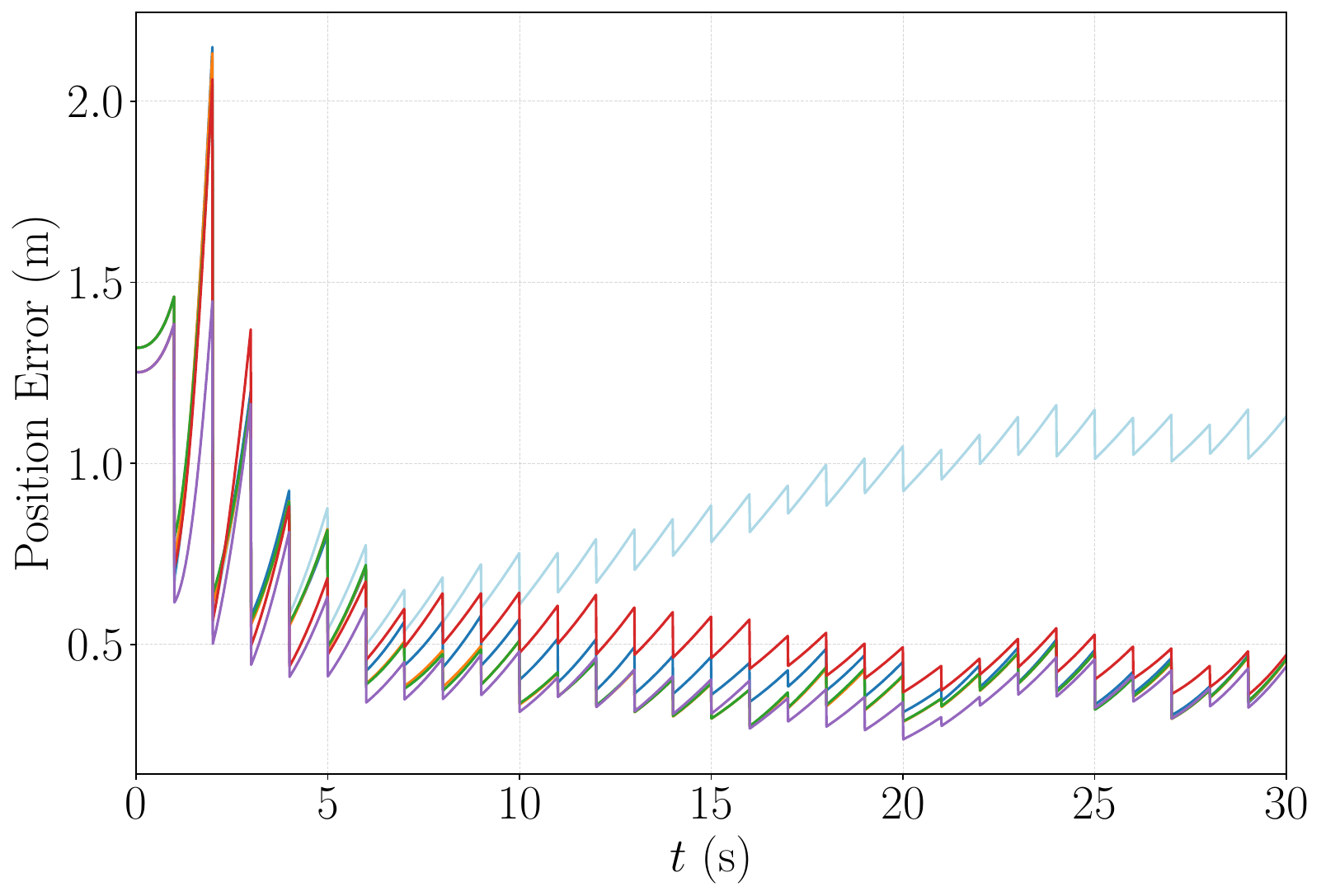}
    \label{fig:inertial_pos}
}
\\
\subfloat{
    \includegraphics[width=0.38\textwidth]{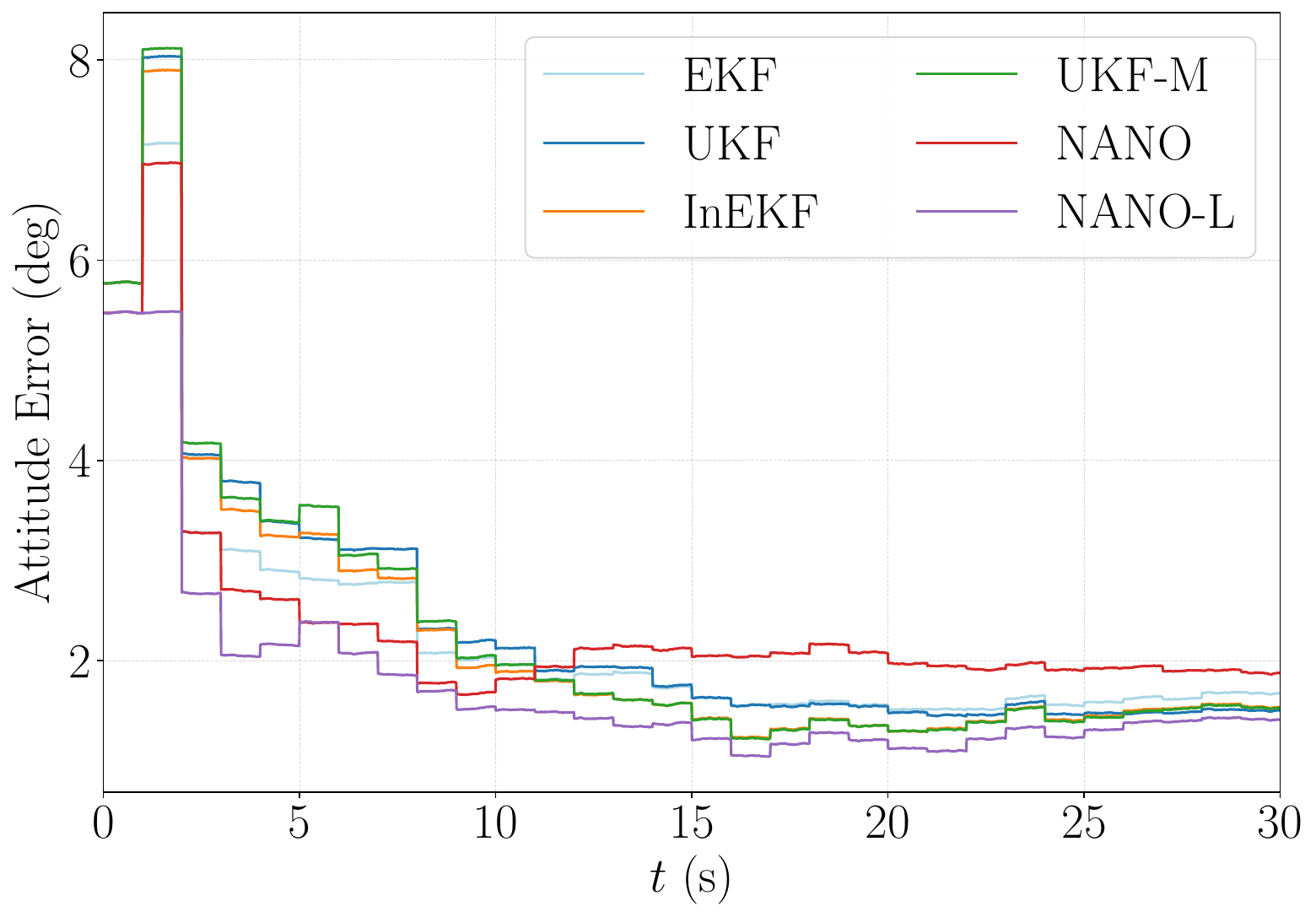}
    \label{fig:inertial_ori}
}
\caption{RMSE for position and orientation at each time step, averaged over 100 Monte Carlo trials.}
\label{fig:inertial}
\end{figure}

The simulation results, presented in Fig.~\ref{fig:inertial_rmse} and Fig.~\ref{fig:inertial}, confirm the superior performance of the proposed NANO-L filter. The aggregated RMSE box plots in Fig.~\ref{fig:inertial_rmse} show that manifold-aware methods significantly outperform their Euclidean counterparts. This confirms the benefit of respecting the underlying Lie group geometry of the state. Among the manifold filters, NANO-L achieves the lowest median error and the most compact distribution for both orientation and position. Its average position estimation error is 18\% lower than NANO's and 16\% lower than UKF-M's. The error-over-time plots in Fig.~\ref{fig:inertial} corroborate this, showing NANO-L maintains a consistently lower error trajectory than other methods. This improved accuracy stems from NANO-L's ability to mitigate the linearization errors inherent in other approaches, validating our filter as a highly accurate and robust solution for state estimation on Lie groups.


\section{Legged Robot Real-World Experiments}
To validate the NANO-L filter beyond the simulated environment, we conducted experiments on a real Unitree GO2 legged robot. This directly tests the filter's performance on the model detailed in Sec.~\ref{sec:modeling}. 

\subsection{Experimental Setup}
\begin{figure}[!t]
\centering
\includegraphics[width=0.4\textwidth]{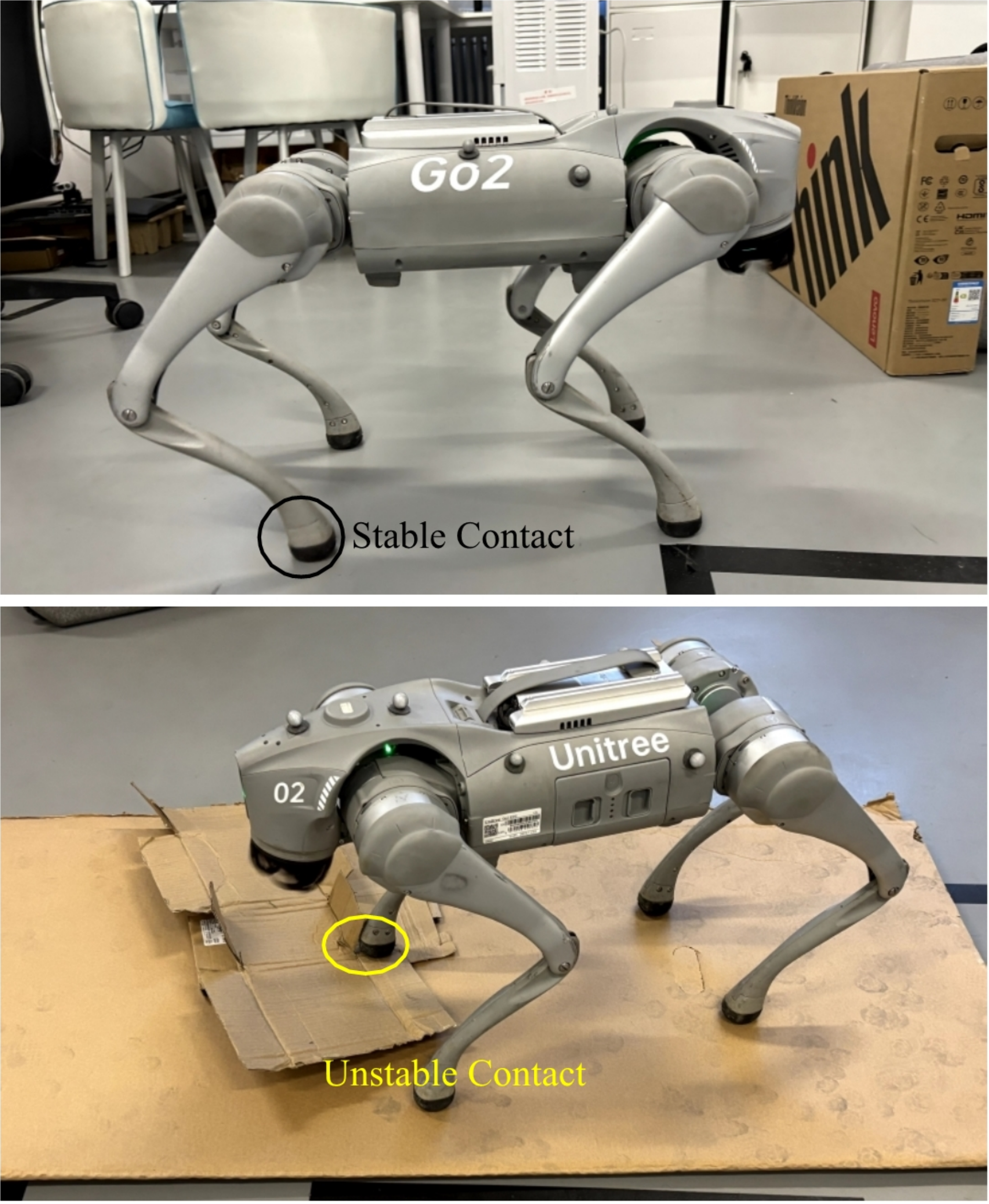}
\caption{Different environments of real-world experiments.}
\label{fig:env}
\end{figure}

A key advantage of legged robots is their ability to adapt to complex terrains \cite{gao2022invariant, sun2024leg}. To evaluate the robustness of the proposed NANO-L filter across different terrains, we designed two distinct experimental environments, as illustrated in Fig.~\ref{fig:env}. These included: a \textit{flat} ground for baseline performance, and an \textit{unstable} surface to challenge the filter with uncertain contact information.

In each environment, the legged robot was controlled to move in a trot gait while sensor data was collected. The onboard sensing suite included one IMU, 12 joint encoders, and four foot contact sensors, all recorded at 200 Hz. In addition, the robot’s ground-truth pose was provided by a high-precision NOKOV motion capture system operating at 100 Hz, as shown in Fig.~\ref{fig:intro}. To minimize the influence of randomness, five 60-second datasets were collected for each environment. For a fair and reproducible comparison, all algorithms were executed on the collected datasets using a laptop equipped with an Intel Core i9-14900HX processor and initialized from the ground-truth state. The noise parameters and algorithm hyperparameters were set consistently for all trials, as detailed in Table~\ref{table.param}.

Finally, in this real-world legged robot experiment, we primarily compare the proposed NANO-L filter against EKF \cite{bloesch2013state, yang2023multi} and InEKF \cite{hartley2020contact, gao2022invariant}, as these filters are widely applied and studied in legged robots. In contrast, UKF-based methods are rarely adopted in practice for such platforms due to their higher computational cost and lower numerical stability. The estimation performance is evaluated using two commonly employed metrics: the absolute trajectory error (ATE) and the relative error (RE) for position, velocity, and orientation, with the specific metric formulations taken from \cite{zhang2018tutorial}. ATE measures the global consistency of the estimated trajectory with respect to the ground truth, while RE reflects the local accuracy by quantifying drift over a fixed time interval (set to 3\,s in our experiments). Together, these two metrics provide complementary insights into both long-term accuracy and short-term consistency of the estimators.

\begin{table}[h]
\label{table.param}
\fontsize{9}{9}\selectfont
\caption{Noise parameters and NANO-L's hyperparameter.}
\centering
\begin{threeparttable}[t]
\renewcommand{\arraystretch}{1.2}
\begin{tabular}{ c c c} 
\toprule[1pt]
Name & Symbol & Value \\
\midrule[0.5pt]
Accelerometer & $\sigma_a$ & $0.2568\ \rm{m/s^2}$  \\
\midrule[0.5pt]
Gyroscope & $\sigma_\omega$ & $0.00139\ \mathrm{rad/s}$  \\
\midrule[0.5pt]
Joint encoder & $\sigma_{e}$ & $0.3\ \rm{rad}$  \\
\midrule[0.5pt]
Slip & $\sigma_{s}$ & $0.001\ \rm{m/s}$  \\
\midrule[0.5pt]
Stopping threshold & $\gamma$ & $10^{-4}$  \\
\bottomrule[1pt]
\end{tabular}
\end{threeparttable}
\end{table}

\subsection{Comparison Results}
\begin{table*}[!t]
\fontsize{8.2}{8.2}\selectfont
\caption{Comparison of ATE, RE, and Computation Time Across Real-World Environments}
\label{table:ate_re}
\centering
\begin{threeparttable}[t]
\renewcommand{\arraystretch}{1.3} 
\begin{tabular}{ c|c| c c c  c c c | c } 
\toprule[1pt]
 \multirow{2} {*}{Environment} & \multirow{2} {*}{Method} & $\rm ATE_{pos}$ & $\rm ATE_{vel}$ & $\rm ATE_{ori}$ & $\rm RE_{pos}$ & $\rm RE_{vel}$ & $\rm RE_{ori}$ & Time\\
 & & $\left [ \rm m \right ]$ & $\left [ \rm m/s \right ]$ & $\left [ \rm rad \right ]$ & $\left [ \rm m \right ]$ & $\left [ \rm m/s \right ]$ & $\left [ \rm rad \right ]$ & $\left [ \rm ms \right ]$ \\
\midrule[0.5pt]
\multirow{3} {*}{Flat} 
& EKF & $0.373\ (0.055)$ & $0.122\ (0.032)$ & $0.020\ (0.005)$ & $0.076\ (0.010)$ & $0.107\ (0.008)$ & $0.017\ (0.005)$ & 3.905\\
& InEKF & $0.281\ (0.099)$ & $0.108\ (0.007)$ & $0.022\ (0.004)$ & $0.072\ (0.010)$ & $0.106\ (0.015)$ & $0.017\ (0.004)$ & \textbf{0.712}\\
\rowcolor{tabGray}
& NANO-L & $\bf 0.212$ $(0.078)$ & $\bf 0.107$ $(0.009)$ & $\bf 0.017$ $(0.005)$ & $\bf 0.065$ $(0.004)$ & $\bf 0.093$ $(0.014)$ & $\bf 0.016$ $(0.003)$ & 3.351\\
\midrule[0.5pt]
\multirow{3} {*}{Unstable} 
& EKF & $0.633\ (0.200)$ & $0.149\ (0.057)$ & $0.021\ (0.002)$ & $0.229\ (0.163)$ & $0.094\ (0.014)$ & $0.016\ (0.003)$ & 3.905\\
& InEKF & $0.401\ (0.108)$ & $0.140\ (0.028)$ & $0.023\ (0.001)$ & $0.162\ (0.082)$ & $0.106\ (0.010)$ & $0.017\ (0.003)$ & \textbf{0.712}\\
\rowcolor{tabGray}
& NANO-L & $\bf 0.236$ $(0.058)$ & $\bf 0.125$ $(0.011)$ & $\bf 0.012$ $(0.001)$ & $\bf 0.104$ $(0.031)$ & $\bf 0.108$ $(0.006)$ & $\bf 0.014$ $(0.004)$ & 3.351\\
\bottomrule[1pt]
\end{tabular}
\end{threeparttable}
\end{table*}

\begin{figure*}[!t]
\centering
\subfloat{
    \includegraphics[width=0.33\textwidth]{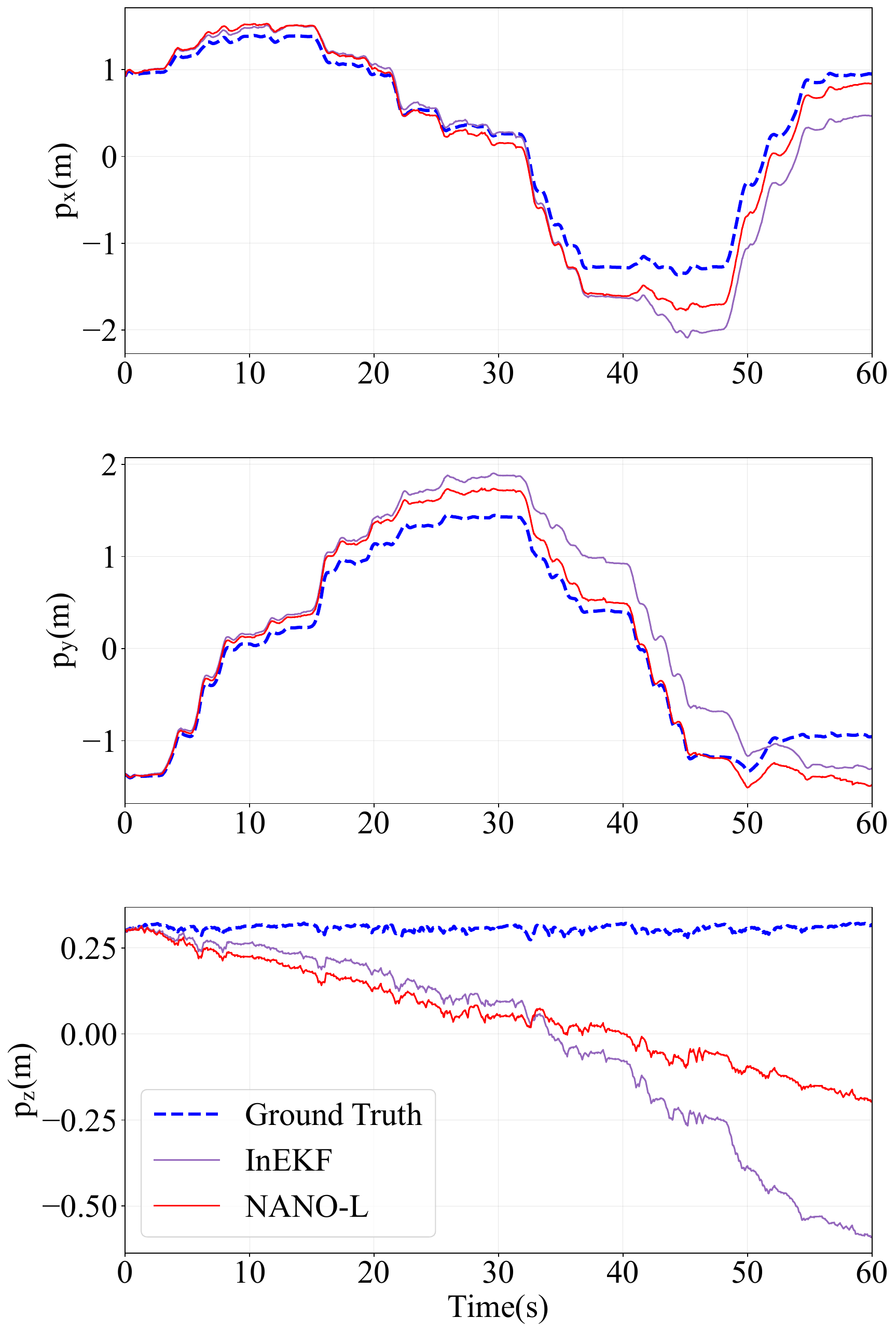}
    \label{fig:unstable_pos}
}
\subfloat{
    \includegraphics[width=0.33\textwidth]{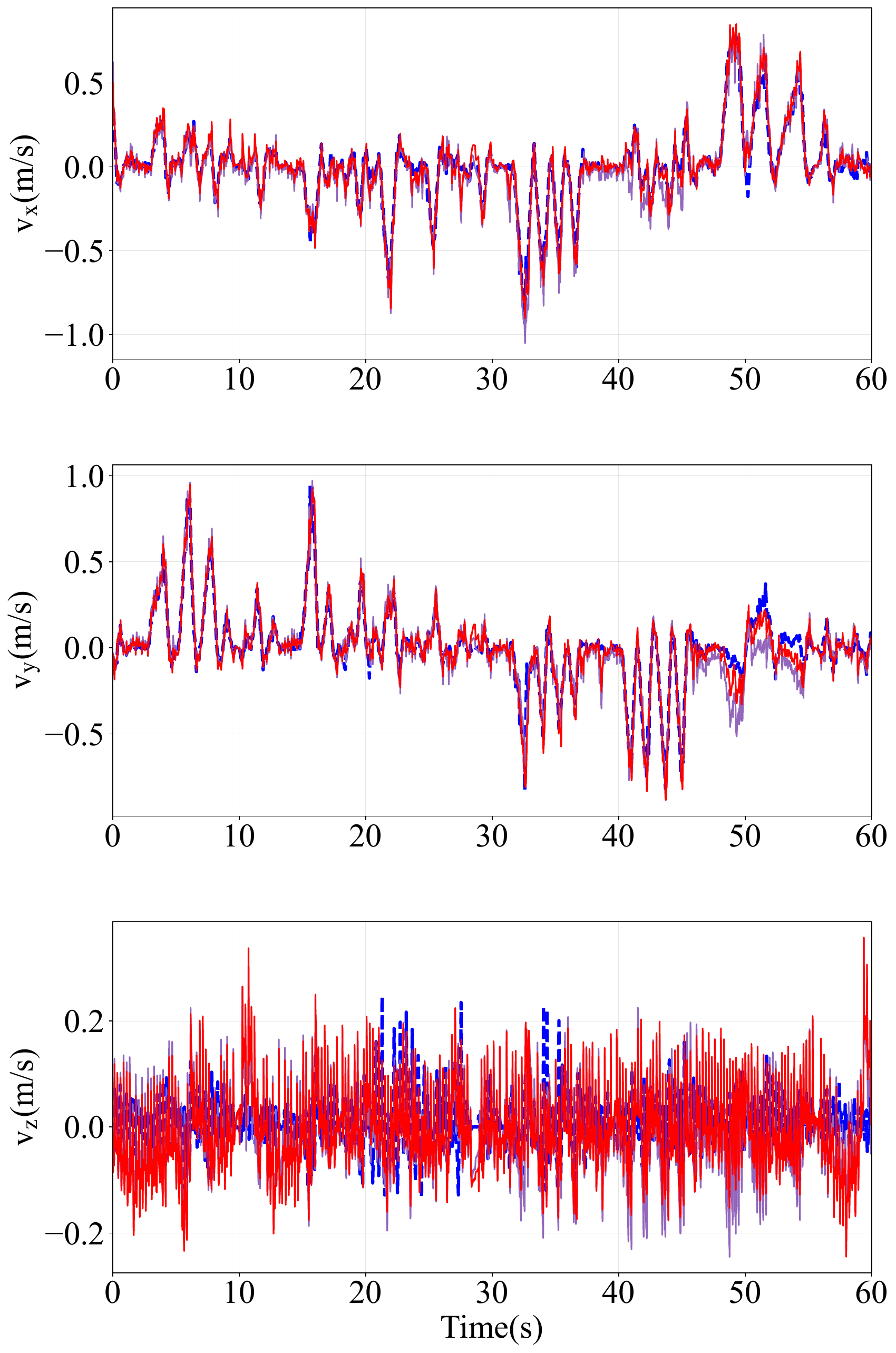}
    \label{fig:unstable_vel}
}
\subfloat{
    \includegraphics[width=0.33\textwidth]{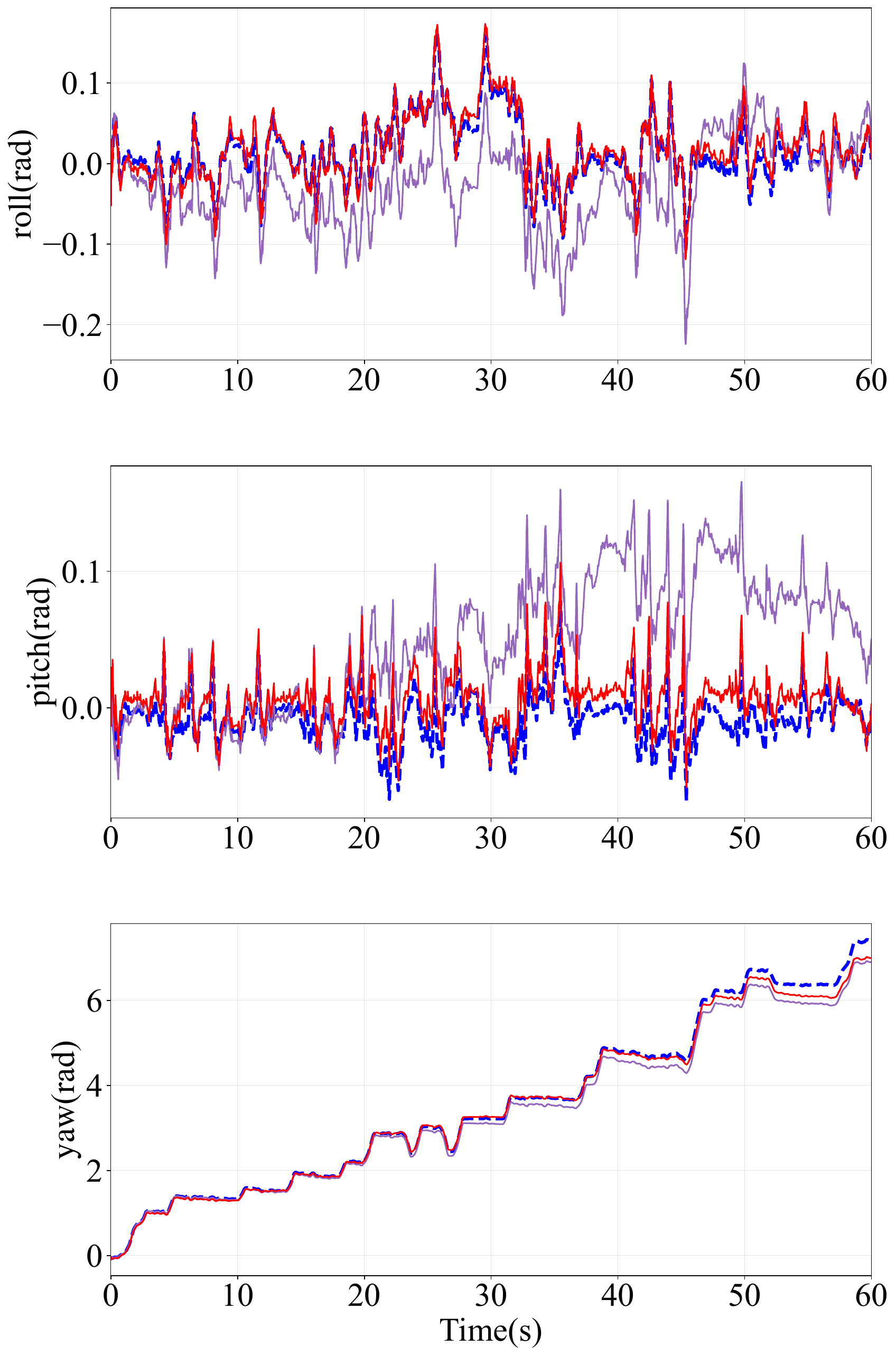}
    \label{fig:unstable_ori}
}
\caption{Estimated position, velocity, and orientation for InEKF and NANO-L on the unstable terrain.}
\label{fig:unstable}
\end{figure*}
Table~\ref{table:ate_re} summarizes the ATE and RE metrics for all filters across the two environments, including the mean (outside the parentheses) and variance (inside the parentheses) of each metric computed from five experimental datasets. Several key observations can be made. First, as expected, the estimation errors for all methods increase on the more challenging unstable terrain, highlighting the difficulty of these scenarios. Second, both manifold-aware filters show a clear advantage over the standard EKF, particularly in reducing position error. Most importantly, our proposed NANO-L filter not only consistently outperforms the state-of-the-art InEKF across nearly all metrics and environments, but also demonstrates superior robustness by showing the least performance degradation on the more challenging terrain. For instance, on the unstable terrain, NANO-L reduces the position ATE by approximately 41\% compared to InEKF. In terms of computational cost, while InEKF is exceptionally fast, NANO-L's processing time of approximately 3.35 ms remains well within the 5 ms budget required for 200 Hz real-time operation.

\begin{figure*}[!t]
\centering
\subfloat{
    \includegraphics[width=0.33\textwidth]{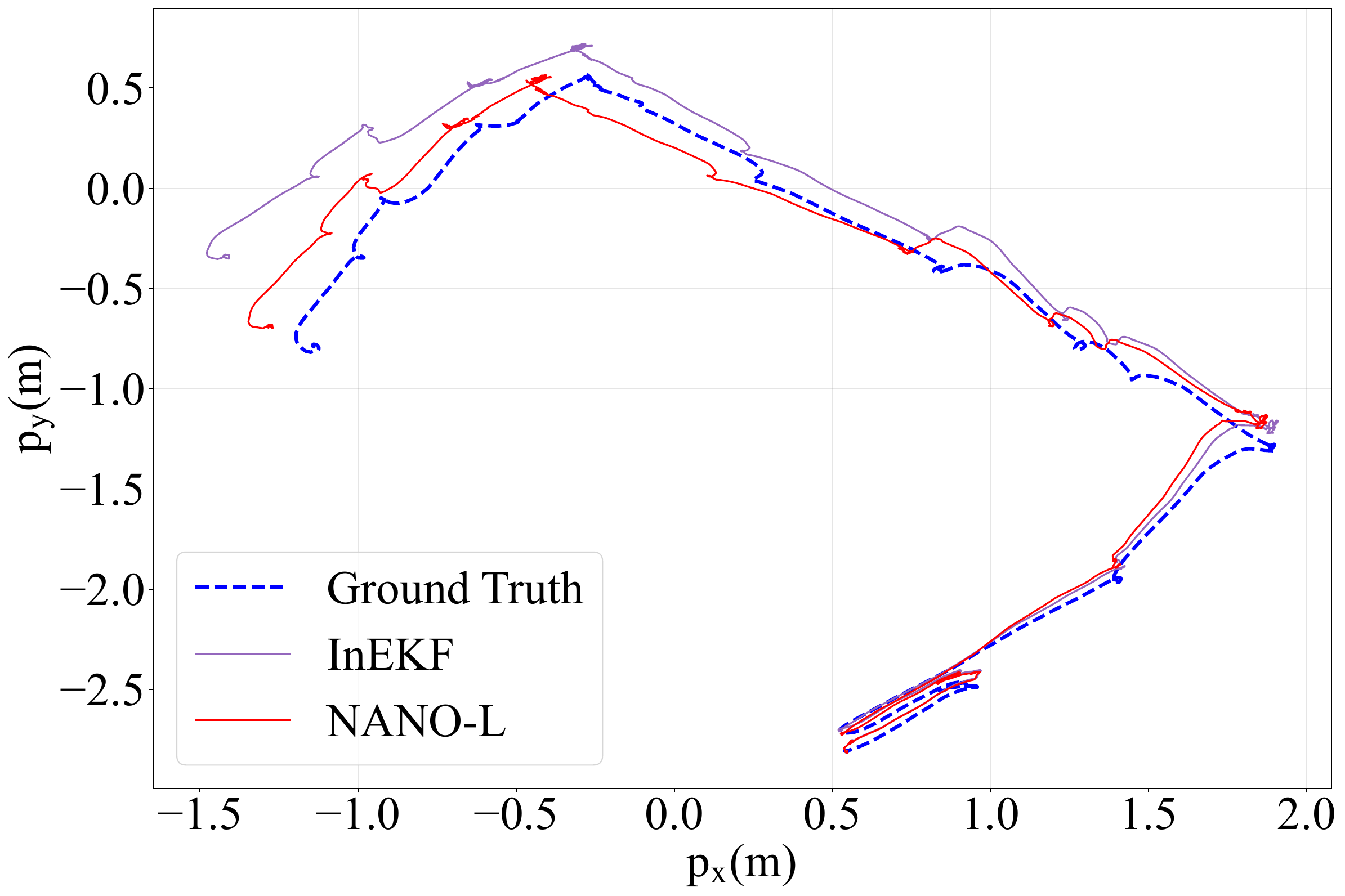}
}
\subfloat{
    \includegraphics[width=0.325\textwidth]{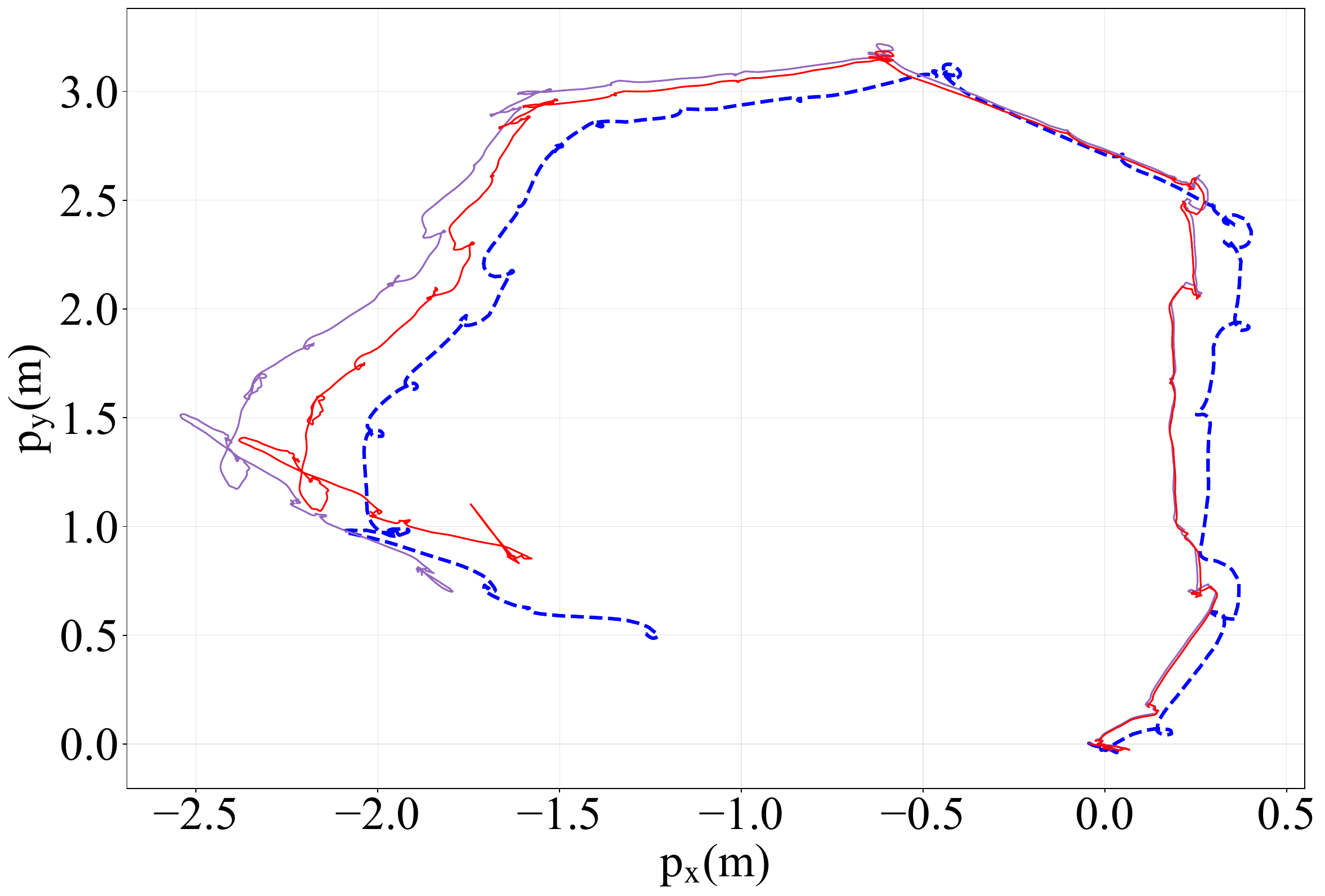}
}
\subfloat{
    \includegraphics[width=0.33\textwidth]{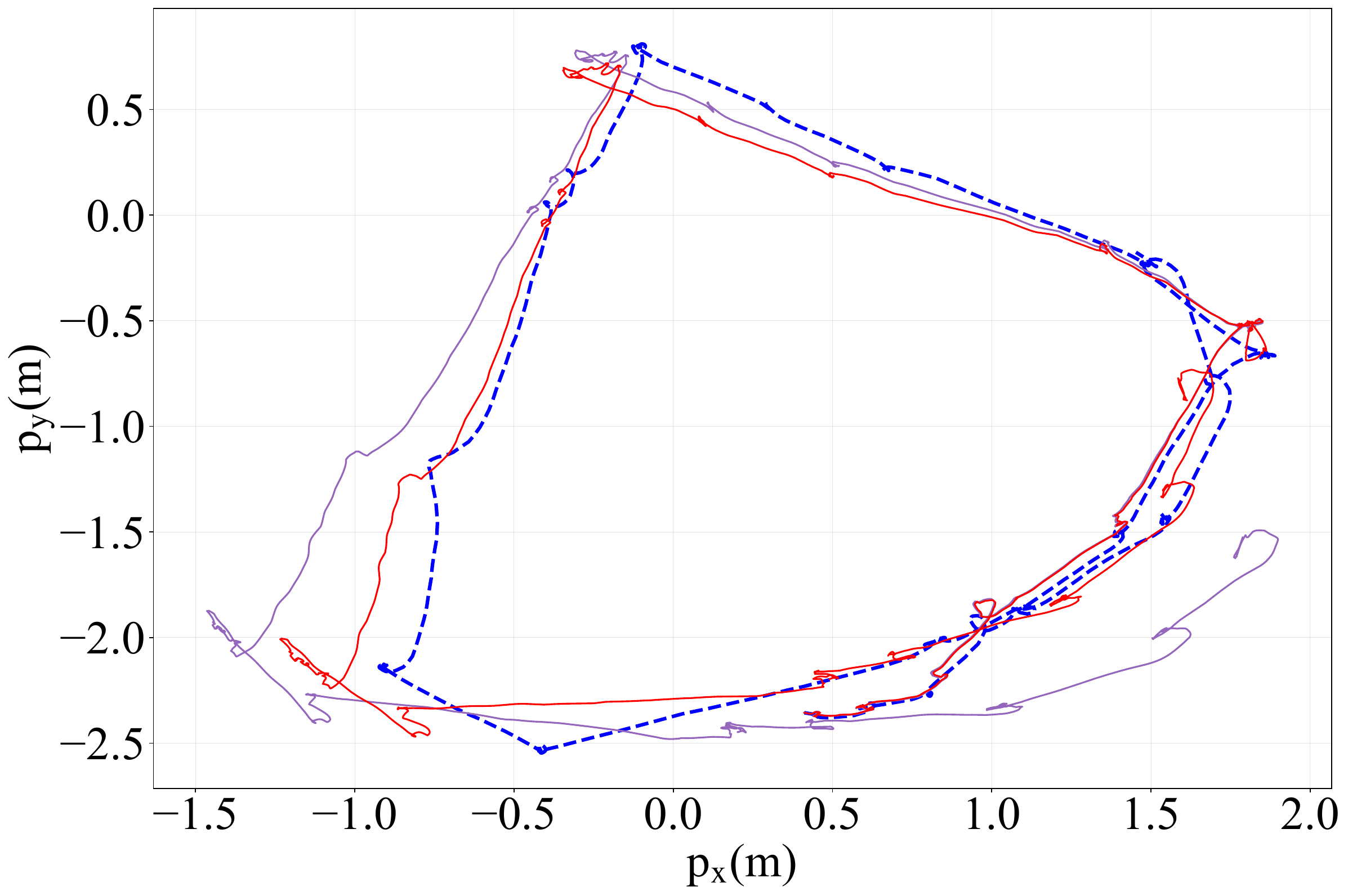}

}
\caption{Estimated trajectories from InEKF and NANO-L on the flat terrain with different distances and durations. From left to right: (9.25 m, 40 s), (10.25 m, 50 s), and (14.6 m, 60 s).}
\label{fig:traj}
\end{figure*}

Fig.~\ref{fig:unstable} provides a visual comparison of the estimated trajectories from a representative trial on the unstable terrain (EKF is omitted for clarity). The plots visually confirm the quantitative findings, showing that the trajectory estimated by NANO-L adheres more closely to the ground truth than that of InEKF across all components. A noteworthy detail in the trajectory plots is the vertical position $p_z$, where all filters exhibit some drift. This is due to the known poor observability of absolute vertical height from leg odometry \cite{gao2022invariant, yoon2023invariant}, though NANO-L still demonstrates a clear tracking advantage. 
Fig.~\ref{fig:traj} presents a comparison of 2D localization from trials conducted on the same flat terrain, where the robot moved at approximately the same speed (\(\approx\)0.2 m/s) over different distances and durations. The results demonstrate that the 2D localization for legged robot accumulates significant drift as travel time and distance increase. In contrast, by avoiding linearization of the observation model, our NANO-L filter effectively mitigates error accumulation, resulting in a notably smaller trajectory drift.

\begin{figure}[!t]
\centering
\includegraphics[width=0.35\textwidth]{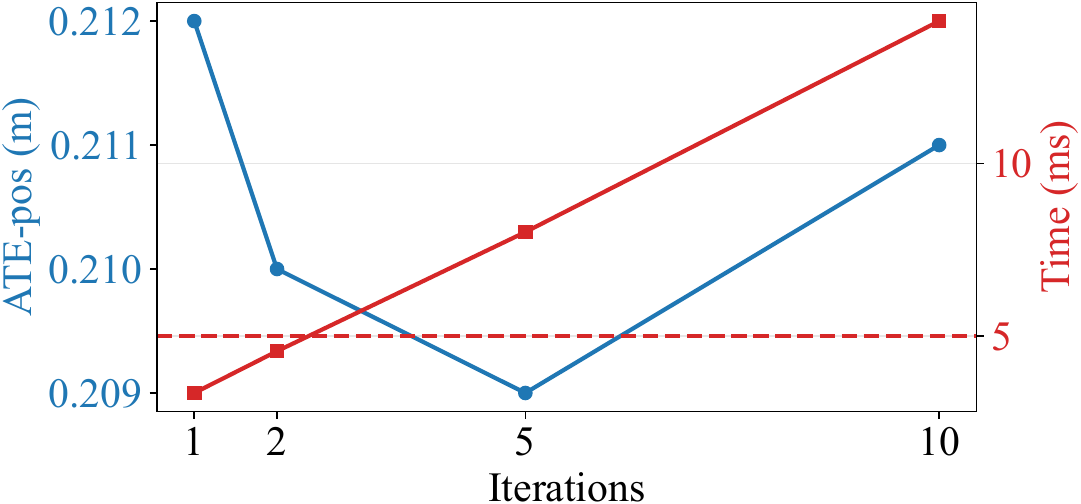}
\caption{Estimation accuracy and computation time of NANO-L on flat terrain with different iteration numbers. The red dashed line indicates the 200 Hz real-time limit.}
\label{fig:ablation}
\end{figure}

Finally, an ablation study on the number of NANO-L iterations shown in Fig.~\ref{fig:ablation} confirms that a single update step provides an excellent trade-off between accuracy and efficiency. This single iteration achieves near-optimal performance while remaining well within the 5 ms computational budget, justifying its use in our comparisons.

\section{Conclusion}
This paper proposes the NANO-L filter, a novel framework for state estimation on Lie group manifolds that eliminates the limitations of local linearization by reformulating manifold filtering as the parameter optimization of a Gaussian-distributed increment in the tangent space. The NANO-L filter solves this optimization using natural gradient descent, which leverages the Fisher information matrix to respect the underlying geometry of the tangent space, resulting in more stable convergence. Furthermore, by exploiting the perturbation model in Lie derivative, we proved that for common invariant observation models, the covariance update admits an exact closed-form solution, eliminating iterative updates and greatly improving efficiency. Real-world experiments on a Unitree GO2 legged robot demonstrated that NANO-L achieves up to 40\% lower estimation error compared to commonly used manifold filters, while maintaining comparable computational cost.

\begin{appendix}
\subsection{Forward Kinematics}
\label{sec:appen}
We denote the joint angles as $\boldsymbol{\phi} = \left[\phi_1; \phi_2; \phi_3\right]$, with $\sin(\phi_i) = s_i$, $\cos(\phi_i) = c_i$ and $\sin(\phi_{i} + \phi_j)=s_{ij}$. The variables $l_h$, $l_t$, and $l_c$ represent the hip length, thigh length, and calf length, respectively. The coordinate of the hip joint in the body frame is denoted as \(\left[o_x,o_y,0\right]\). The forward kinematics function can then be written as
\begin{equation*}
\boldsymbol{fk}(\boldsymbol{\phi}) = \begin{bmatrix}
o_x \\
o_y \\
0
\end{bmatrix} + \begin{bmatrix}
-l_t s_2 - l_c s_{23} \\
l_h c_1 + l_t c_2 s_1 + l_c s_1 c_{23} \\
l_h s_1 - l_t c_1 c_2 - l_c c_1 c_{23}
\end{bmatrix}.
\end{equation*}
Differentiate $\boldsymbol{fk}(\boldsymbol{\phi})$ with respect to $\boldsymbol{\phi}$ to obtain the kinematics Jacobian matrix:
\begin{equation*}
\begin{aligned}
&\mathbf{J}_{\text{fk}}(\boldsymbol{\phi})  \\ =
&\begin{bmatrix}
0 & -l_c c_{23} - l_t c_2 & -l_c c_{23}\\
l_t c_1 c_2 - l_h s_1 + l_c c_1 c_{23} & -s_1(l_c s_{23} + l_t s_2) & -l_c s_{23} s_1 \\
l_t c_2 s_1 + l_h c_1 + l_c s_1 c_{23} & c_1(l_c s_{23} + l_t s_2) & l_c s_{23}c_1
\end{bmatrix}.
\end{aligned}
\end{equation*}

\subsection{Derivation of the Analytic Derivatives}
\label{sec:appen_b}
For simplicity, we first study the derivative of \(\mathrm{Exp}(\delta\bm{\xi})\bm{X}^{-1}\bm{b}\) with respect to a small perturbation \(\delta\bm{\xi}=\left[\delta \bm{\phi} ; \delta \bm{v} ; \delta \bm{p} ; \delta \bm{s}\right]
\) near zero, yielding
\begin{align*}
&\frac{\partial \mathrm{Exp}(\delta\bm{\xi})\bm{X}^{-1}\bm{b}}{\partial \delta\bm{\xi}} \\
&= \lim_{\delta\bm{\xi}\rightarrow \bm{0}} \frac{\mathrm{Exp}(\delta\bm{\xi})\bm{X}^{-1}\bm{b} - \bm{X}^{-1}\bm{b}}{\delta\bm{\xi}} \\
&= \lim_{\delta\bm{\xi}\rightarrow \bm{0}} \frac{(\bm{I}+\delta\bm{\xi}^\wedge)\bm{X}^{-1}\bm{b} - \bm{X}^{-1}\bm{b}}{\delta\bm{\xi}} \\
&=\lim_{\delta\bm{\xi}\rightarrow \bm{0}} \frac{\delta\bm{\xi}^\wedge\bm{X}^{-1}\bm{b}}{\delta\bm{\xi}} \\
&= \lim_{\delta\bm{\xi}\rightarrow \bm{0}} \frac{\begin{bmatrix}
\delta \bm{\phi} ^\wedge & \delta \bm{v} & \delta \bm{p} & \delta \bm{s} \\
\bm{0}_{3\times 3} & \bm{0}_{3\times 1} & \bm{0}_{3\times 1} & \bm{0}_{3\times 1}
\end{bmatrix} \begin{bmatrix}
\bm{R}^\top(\bm{s} - \bm{p}) \\
0 \\
1 \\
-1
\end{bmatrix}}{\left[\delta \bm{\phi} ; \delta \bm{v} ; \delta \bm{p} ; \delta \bm{s}\right]}\\
&= \lim_{\delta\bm{\xi}\rightarrow \bm{0}} \frac{\begin{bmatrix}
\big(\bm{R}(\bm{p} - \bm{s})\big)^\wedge \delta \bm{\phi} + \delta \bm{p} - \delta \bm{s} \\
0 \\
0 \\
0
\end{bmatrix}}{\left[\delta \bm{\phi} ; \delta \bm{v} ; \delta \bm{p} ; \delta \bm{s}\right]}\\
&=\begin{bmatrix}
\big(\bm{R}(\bm{p} - \bm{s})\big)^\wedge & \bm{0}_{3 \times 3} & \bm{\mathrm{I}}_{3\times 3} & -\bm{\mathrm{I}}_{3\times 3} \\
\bm{0}_{3 \times 3} & \bm{0}_{3 \times 3} & \bm{0}_{3 \times 3} & \bm{0}_{3 \times 3}.
\end{bmatrix}^\top := (\bm{X}^{-1}\bm{b})^\odot.
\end{align*}
The first equality in this formula applies the  perturbation model on Lie derivatives \cite{barfoot2024state}. In our NANO-L filter, the increment \(\bm{\xi}_t\) is such a small quantity close to zero; therefore, we have
\begin{equation}
\nonumber
\frac{\partial \mathrm{Exp}(-\bm{\xi}_t)\hat{\bm{X}}_{t|t-1}^{-1}\bm{b}}{\partial \bm{\xi}_t} = -(\hat{\bm{X}}_{t|t-1}^{-1}\bm{b})^\odot.
\end{equation}
Finally, we can derive the first-order and second-order derivatives of the negative log-likelihood \(\ell(\bm{\xi}_t, \bm{y}_t)\) with respect to \(\bm{\xi}_t\) as
\begin{equation}
\nonumber
\begin{aligned}
&\frac{\partial \ell(\bm{\xi}_t, \bm{y}_t)}{\partial \bm{\xi}_t}\\
&=\frac{\partial (\bm{y}_t - \mathrm{Exp}(-\bm{\xi}_t)\hat{\bm{X}}_{t|t-1}^{-1}\bm{b})}{\partial \bm{\xi}_t}\frac{\partial \ell(\bm{\xi}_t, \bm{y}_t)}{\partial (\bm{y}_t - \mathrm{Exp}(-\bm{\xi}_t)\hat{\bm{X}}_{t|t-1}^{-1}\bm{b})}\\
&=(\hat{\bm{X}}_{t|t-1}^{-1}\bm{b})^\odot\bm{\Gamma}_t^{-1}(\bm{y}_t - \mathrm{Exp}(-\bm{\xi}_t)\hat{\bm{X}}_{t|t-1}^{-1}\bm{b}),\\
&\frac{\partial^2 \ell(\bm{\xi}_t, \bm{y}_t)}{\partial\bm{\xi}_t^\top\partial \bm{\xi}_t} \\
&= \frac{\partial (\hat{\bm{X}}_{t|t-1}^{-1}\bm{b})^\odot \bm{\Gamma}_t^{-1} \bm{y}_t}{\partial \bm{\xi}_t^\top} - (\hat{\bm{X}}_{t|t-1}^{-1}\bm{b})^\odot \bm{\Gamma}_t^{-1}\frac{\partial \mathrm{Exp}(-\bm{\xi}_t)\hat{\bm{X}}_{t|t-1}^{-1}\bm{b}}{\partial \bm{\xi}_t ^\top} \\
&={(\hat{\bm{X}}_{t|t-1}^{-1}\bm{b})^\odot} \bm{\Gamma}_t^{-1} {(\hat{\bm{X}}_{t|t-1}^{-1}\bm{b})^\odot}^\top.
\end{aligned}
\end{equation}

\subsection{System Model in the Numerical Simulation}
\label{sec:appen_c}
The state of the vision-aided inertial navigation system in the numerical simulation is defined as
\begin{equation}
\nonumber
\bm{X}_t = 
\begin{bmatrix}
\bm{R}_t & \bm{v}_t & \bm{p}_t  \\
\bm{0}_{3\times3} & \bm{0}_{3\times1} & \bm{0}_{3\times1} 
\end{bmatrix} 
\in \mathrm{SE}_2(3).
\end{equation}
The state propagation model is given by
\begin{equation}
\nonumber
\begin{aligned}
\frac{\mathrm{d}}{\mathrm{d}t}\bm{X}_t &= \bm{f}_{\bm{u}_t}(\bm{X}_t) + \bm{X}_t\bm{n}_t^\wedge \\
&= 
\begin{bmatrix}
\bm{R}_t \tilde{\bm{\omega}}_t^\wedge  & \bm{R}_t \tilde{\bm{a}}_t + \bm{g} & \bm{v}_t \\
\bm{0}_{3\times3} & \bm{0}_{3\times1} & \bm{0}_{3\times1} 
\end{bmatrix} 
+ \bm{X}_t\bm{n}_t^\wedge,
\end{aligned}
\end{equation}
and the observation model can be expressed as
\begin{equation}
\nonumber
\begin{aligned}
\underbrace{\begin{bmatrix}
\bm{c} \\
0 \\
1 \\
\end{bmatrix}}_{\bm{y}_t} 
= 
\bm{X}_t^{-1}
\underbrace{\begin{bmatrix}
  \bm{m} \\
  0 \\
  1 \\
\end{bmatrix}}_{\bm{b}} 
+ 
\underbrace{\begin{bmatrix}
  \bm{\eta}^c \\
  0 \\
  0 \\
\end{bmatrix}}_{\bm{\eta}_t},   
\end{aligned}
\end{equation}
where \(\bm{c} \in \mathbb{R}^3\) represents the position of the landmark \(\bm{m} \in \mathbb{R}^3\) as observed by the camera, and \(\bm{\eta}^c \in \mathbb{R}^3\) denotes zero-mean Gaussian noise.

\end{appendix}

\bibliographystyle{IEEEtran}
\bibliography{ref}

\end{document}